\documentclass[journal]{IEEEtran}
\usepackage[absolute,overlay]{textpos}
\usepackage{tikz}
\usepackage{amsfonts}
\usepackage{diagbox}
\usepackage{bbding}
\usepackage{times}
\usepackage{soul}
\usepackage{url}
\usepackage[hidelinks]{hyperref}
\usepackage[utf8]{inputenc}
\usepackage{graphicx}
\usepackage{amsmath}
\usepackage{amsthm}
\usepackage{booktabs}
\usepackage{multirow}
\usepackage{mathtools}
\usepackage{algorithmic}
\usepackage{flushend}
\usepackage[numbers]{natbib}
\usepackage{subfigure}
\usepackage[misc]{ifsym}

\usepackage{lipsum}

\usepackage {tcolorbox}

\usepackage[linesnumbered, ruled,vlined]{algorithm2e}
\SetKwInput{KwInput}{Input}                
\SetKwInput{KwOutput}{Output}

\newcommand{\ie}{i.e.}
\newcommand{\eg}{e.g.}
\usepackage[switch,pagewise]{lineno}
\usepackage{lipsum}

\title{Hierarchical Weight Averaging for \\Deep Neural Networks}

\author{Xiaozhe~Gu$^*$,~Zixun Zhang$^*$,~Yuncheng Jiang,~Tao Luo,~Ruimao Zhang,~Shuguang Cui, Zhen Li$^\textrm{\Letter}$
\IEEEcompsocitemizethanks{ \IEEEcompsocthanksitem  
This work was supported in part by National Key R\&D program of China with grant No.2018YFB1800800, by the Basic Research Project No. HZQB-KCZYZ-2021067 of Hetao Shenzhen HK S\&T Cooperation Zone, by Shenzhen-Hong Kong Joint Funding No. SGDX20211123112401002, by Shenzhen General Program No. JCYJ20220530143600001, by Guangdong Research Project No. 2017ZT07X152 and No. 2019CX01X104, by the Guangdong Provincial Key Laboratory of Future Networks of Intelligence (Grant No. 2022B1212010001), by the Guangdong Provincial Key Laboratory of Big Data Computing, The Chinese University of Hong Kong, Shenzhen, by the NSFC 61931024\&8192 2046, by Zelixir biotechnology company Fund, by Tencent Open Fund, by the Singapore Government's Research, Innovation and Enterprise 2020 Plan (Advanced Manufacturing and Engineering domain) under Grant A1892b0026.
\\
X. Gu,  Z. Zhang, Y. Jiang, S. Cui, and Z. Li are with the Future Network of Intelligence Institute (FNii), the School of Science and Engineering (SSE), and the Guangdong Provincial Key Laboratory of Future Networks of Intelligence, and the Chinese University of Hong Kong, Shenzhen (CUHKSZ). And S. Cui is also with the Pengcheng Laboratory, Shenzhen.
R. Zhang is with the School of Data Science (SDS), the Chinese University of Hong Kong, Shenzhen.
T. Luo is with the Institute of High Performance Computing, Agency for Science, Technology and Research (A*STAR), Singapore. 
(E-mail: guxi0002@e.ntu.edu.sg, zixunzhang@link.cuhk.edu.cn, yunchengjiang@link.cuhk.edu.cn,  leto.luo@gmail.com, zhangruimao@cuhk.edu.cn, shuguangcui@cuhk.edu.cn, lizhen@cuhk.edu.cn)\\
$^*$: These authors contributed equally to this work.\\
$^\textrm{\Letter}$: Corresponding Author.
}
}

\begin{document}

\maketitle

\begin{abstract}
Despite the simplicity, stochastic gradient descent (SGD)-like algorithms are successful in training deep neural networks (DNNs). 
Among various attempts to improve SGD, weight averaging (WA), which averages the weights of multiple models,  has recently received much attention in the literature. 
Broadly, WA falls into two categories: 1) online WA, which averages  the weights of multiple models trained in parallel, is designed for reducing the gradient communication overhead of parallel mini-batch SGD, and 2) offline WA, which  averages  the weights of one model at different checkpoints, is typically used to improve the generalization ability of  DNNs.   
Though online and offline WA are  similar in form, they are seldom associated with each other.   { Besides, these methods typically perform either offline parameter averaging or online parameter averaging, but not both.} In this work, we firstly attempt  to incorporate   online and offline WA into  a general training framework termed Hierarchical Weight Averaging (HWA). 
By leveraging both the online and offline averaging manners, HWA  is able to achieve both faster convergence speed and superior generalization performance without any fancy learning rate adjustment.   {Besides, we also analyze the issues faced by existing WA methods, and how our HWA address them, empirically.}
Finally, extensive experiments verify that HWA outperforms the state-of-the-art  methods significantly.

\end{abstract}


\section{Introduction}
\label{sec:intro}

\IEEEPARstart{D}{eep} neural networks   have evolved to the critical technique for a  variety of computer vision problems, such as image classification~\cite{krizhevsky2012imagenet,krizhevsky2009learning}, object detection~\cite{donahue2014decaf}, semantic segmentation~\cite{long2015fully} and point cloud classification~\cite{hackel2017semantic3d} since the breakthrough made in the ILSVRC competition~\cite{krizhevsky2012imagenet}. Typically, DNNs are trained with variants of stochastic gradient descent (SGD) algorithms, in conjunction with a decaying learning rate.  Various techniques such as adaptive learning
rate schemes~\cite{duchi2011adaptive} and advanced  learning rate scheduling strategies~\cite{loshchilov2016sgdr,smith2017cyclical} have been developed to improve SGD.    Among these attempts to improve SGD, Weight Averaging (WA) is a simple but effective one and can be broadly categorized into two types: 1) online WA~\cite{lin2018don,mcmahan2017communication,mcdonald2009efficient,zhang2016parallel,stich2018local,zinkevich2010parallelized}, which is designed for reducing the gradient communication overhead of parallel mini-batch SGD and 2) offline WA~\cite{izmailov2018averaging}, which is  typically used to improve the  generalization ability of DNNs. 

\begin{figure}[t!]
\centering
\subfigure[Online Weight Averaging]{
        \begin{minipage}[t]{0.45\textwidth}
        \includegraphics[width=\textwidth]{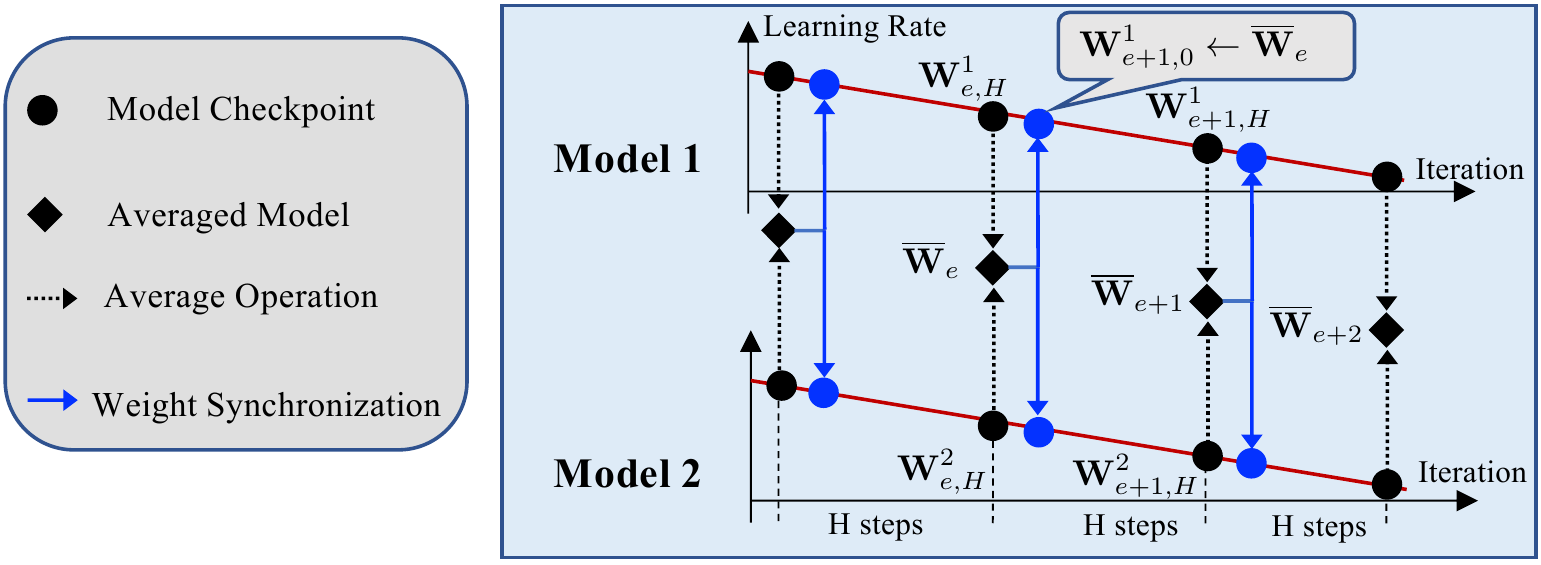}
        \end{minipage}
}
\centering
\centering
\subfigure[Offline Weight Averaging]{
        \begin{minipage}[t]{0.45\textwidth}
        \includegraphics[width=\textwidth]{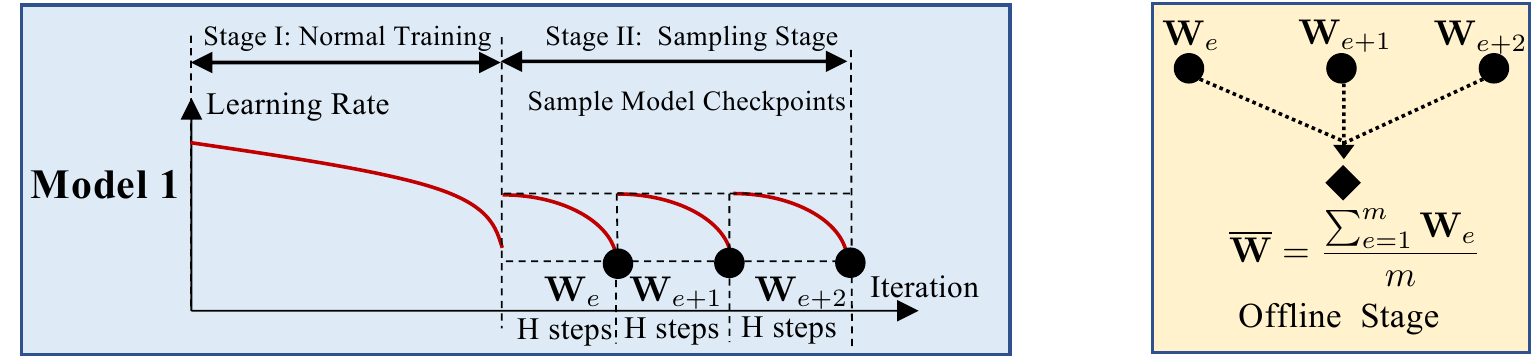}
      \end{minipage}
}
\subfigure[Proposed Hierarchical Weight Averaging]{
        \begin{minipage}[t]{0.45\textwidth}
        \includegraphics[width=\textwidth]{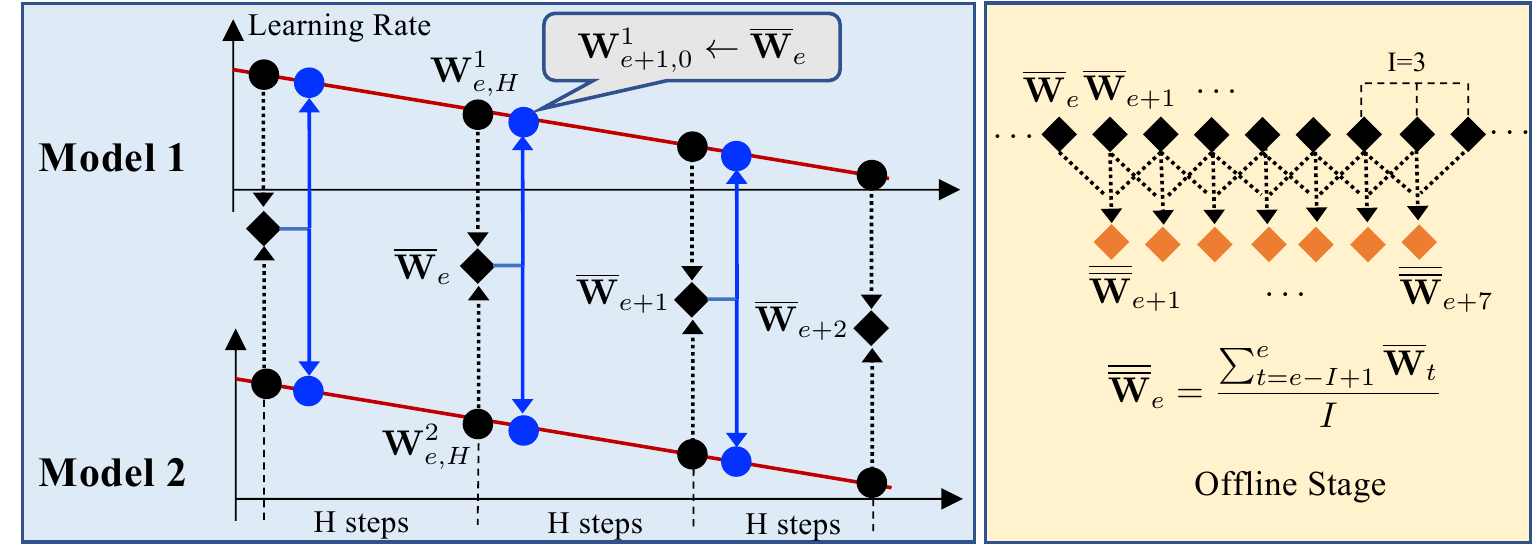}
      \end{minipage}
}

\caption{(a) Online WA  periodically averages the weights of individual models in parallel $\overline{\mathbf W}_{e}=\frac{\sum_{k=1}^K\mathbf W_{e,H}^k }{K}$ and then updates the weights of each model to the averaged weights $\mathbf W_{e+1,0}^k\leftarrow \overline{\mathbf W}_{e}$. (b) Offline WA  averages  multiple points along the trajectory of SGD with a cyclical or  large constant learning rate, and then takes the average. (c)  HWA  leverages both online and offline WA. At the online stage, it periodically synchronizes the weights of individual models trained in parallel with the averaged weights.  At the offline stage, it also further averages the averaged solution $\overline{\mathbf W}_{e}$ sampled at different synchronization cycles with a slide window of length $I$, \ie,  $\overline{\overline{\mathbf W}}_{e}=\frac{\sum_{t=e-I+1}^{e}}{I} \overline{\mathbf W}_{t}$.}
\label{fig:L123}
\end{figure}

As illustrated in Figure~\ref{fig:L123}(a), online WA  averages multiple models trained in parallel every $H$ iterations and then synchronizes  the weights of each model to the averaged weights periodically. The interval  between two  parameter synchronization operations can be named as a  \emph{synchronization cycle} and $H$ is the corresponding \emph{synchronization period}.  The procedure of online WA  is  similar to parallel mini-batch SGD~\cite{li2014communication}, which takes the average of  gradients from multiple models  in parallel, and then  updates each local model using an SGD update with the averaged gradient. 
In other words, at every iteration, parallel mini-batch SGD can be interpreted as  updating each local model with  a single SGD step and then replacing each model with the averaged solution.  Since  parallel mini-batch SGD requires exchanging of local gradient information among all models at every iteration,  the corresponding heavy communication overhead becomes one of its major issues.  
Unlike parallel mini-batch SGD, online WA has significantly less communication overhead as it evolves each  model for $H$ (\eg, $H\in [1,64]$ in~\cite{lin2018don}, $H=10$ in~\cite{zhang2016parallel}, $H\in [4,32]$ in ~\cite{yu2019parallel}) sequential SGD updates before  parameter synchronization. While online WA can speed up the training process, it does not perform very differently in terms of generalization performance.

Unlike online WA, which periodically synchronizes the weights of each model  to the averaged weights, offline WA (\ie, Stochastic Weight Averaging (SWA)~\cite{izmailov2018averaging})  does not participate in the training process directly, and the averaged weights are invisible to the optimizer.   As illustrated in Figure~\ref{fig:L123}(a),   offline WA divides the training process into two stages, \ie,  Stage I: normal training stage   and  Stage II: sampling stage. 
While there is no difference between offline WA and standard SGD at Stage I, offline WA samples  a model checkpoint every $H$ iterations periodically   with \emph{a cyclical or large constant learning rate scheduler}~\cite{loshchilov2016sgdr,smith2017cyclical}   at Stage II, and then takes the average of the sampled model checkpoints to obtain the averaged model. 

Unlike online WA, offline WA evolves a model for more iterations (\eg, $H=1000$) before a model checkpoint is sampled.  While standard SGD typically converges to
a point near the boundary of local minima,  offline WA is able to find a point centered in this region with better generalization performance. However, it does not exhibit better  performance in terms of training efficiency than mini-batch SGD.

\begin{figure}[t!]
\centering
\includegraphics[width=0.4\textwidth]{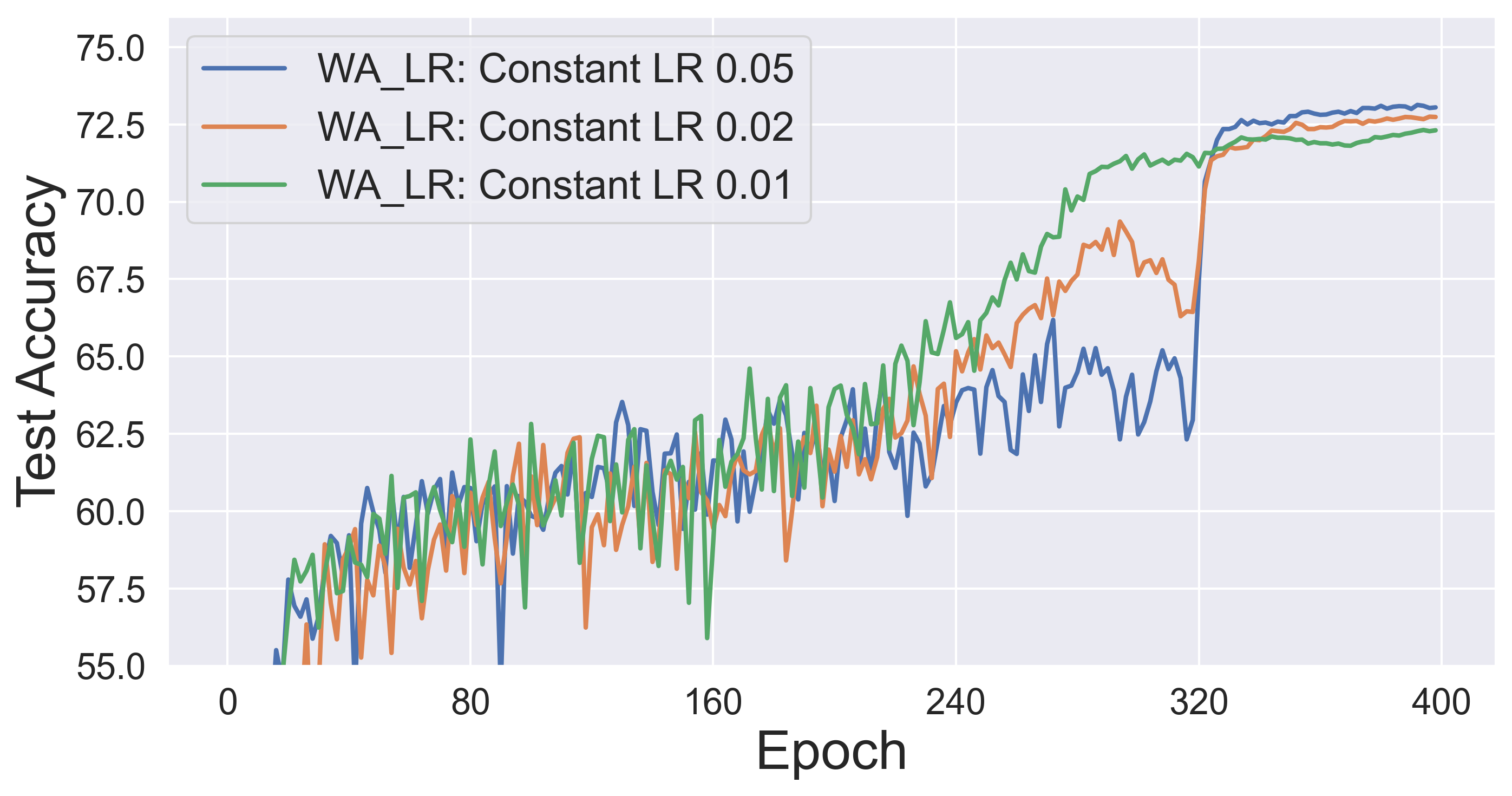}

\caption{
The  top-1  test accuracy of offline WA   on CIFAR100 for  ResNet56 with different constant learning rates  ($\{0.05,0.02,0.01\}$) at Stage II ($321-400$ epochs).  
}
\label{fig:VS}
\end{figure}

Since offline WA performs averaging at the end of the training process, it requires choosing when to start sampling model checkpoints. A choice that is either too early or too late  can be detrimental to  the generalization performance of the averaged model. Besides, offline WA  uses  \emph{a cyclical or large constant learning rate scheduler} (WA\_LR) to sample model checkpoints  at the sampling stage. Since the learning rate  has a  significant impact on the locations of the sampled model checkpoints on the loss surface, the performance of offline WA is sensitive to the choice of  WA\_LR.   As shown in Figure~\ref{fig:VS},    the  test accuracy  of offline WA  varies greatly  when  different constant learning rates  are used at Stage II.   With an improper setting,  offline WA may even have a negative impact on the generalization performance (\eg, ResNet110 on CIFAR100 in  Table~\ref{tab:cifarcomp}). 

To address the issues of offline WA, and combine the advantages of  the two types of WA,  in this work, we propose a novel weight averaging paradigm named Hierarchical Weight Averaging (HWA) to combine  \textit{online WA} and \textit{offline WA} within a unified training framework. As illustrated in Figure~\ref{fig:L123} (c), HWA performs averaging in both online stage and offline stage. During  the online stage, HWA trains multiple models in parallel   with a normal decaying learning rate scheduler. After each model is evolved for  $H$ iterations,  HWA  synchronizes the weights of each model to the averaged weights  $\mathbf  W_{e+1,1}^k\leftarrow \overline{\mathbf W}_{e}$, and then restarts training.

However, unlike the  existing high-frequency online WA, we propose to use  \textbf{\emph{low-frequency online WA}} for HWA, which performs the parameter  synchronization operation with a much lower frequency (\ie, larger synchronization period  $H$). In practice, we observe that the low-frequency online WA  simulates the `restart mechanism' (\eg ~\cite{loshchilov2016sgdr}) for DNN training  and  improves generalization performance.

Later at the offline stage, HWA further  takes the average of   $\overline{\mathbf W}_{e}$ at different synchronization cycles with  a slide window of length $I$, \ie,  $\overline{\overline{\mathbf W}}_{e}=\frac{\sum_{t=e-I+1}^{e}}{I} \overline{\mathbf W}_{t}$.   Unlike existing offline WA which requires a specific learning rate scheduler and takes the average of all the sampled model checkpoints,  our proposed \textbf{\emph{slide-window based offline WA}} only averages the sampled model checkpoints  within the slide window and works well with a regular learning rate scheduler (\eg, a cosine decay of learning rate over the entire budget~\cite{loshchilov2016sgdr}).

\begin{table}[h!]
\footnotesize
  \centering
  \begin{tabular}{|l|c|c|}
\hline
~& $\uparrow$ Training Efficiency&  $\uparrow $ Generalization Performance\\
\hline 
a: Offline WA &\XSolidBrush&\Checkmark\\
b: Online WA&\Checkmark&\XSolidBrush\\
c: HWA&\Checkmark&\Checkmark\\
Rank&  c$>$b$>$a &    c$>$a$>$b \\
\hline
  \end{tabular}
  \vspace{3pt}
  \caption{Comparison of Online WA, Offline WA and HWA in terms of improving training efficiency and   generalization performance }
  \label{tab:tbcompare}
\end{table}

\begin{figure}[t!]
\centering
\includegraphics[width=0.4\textwidth]{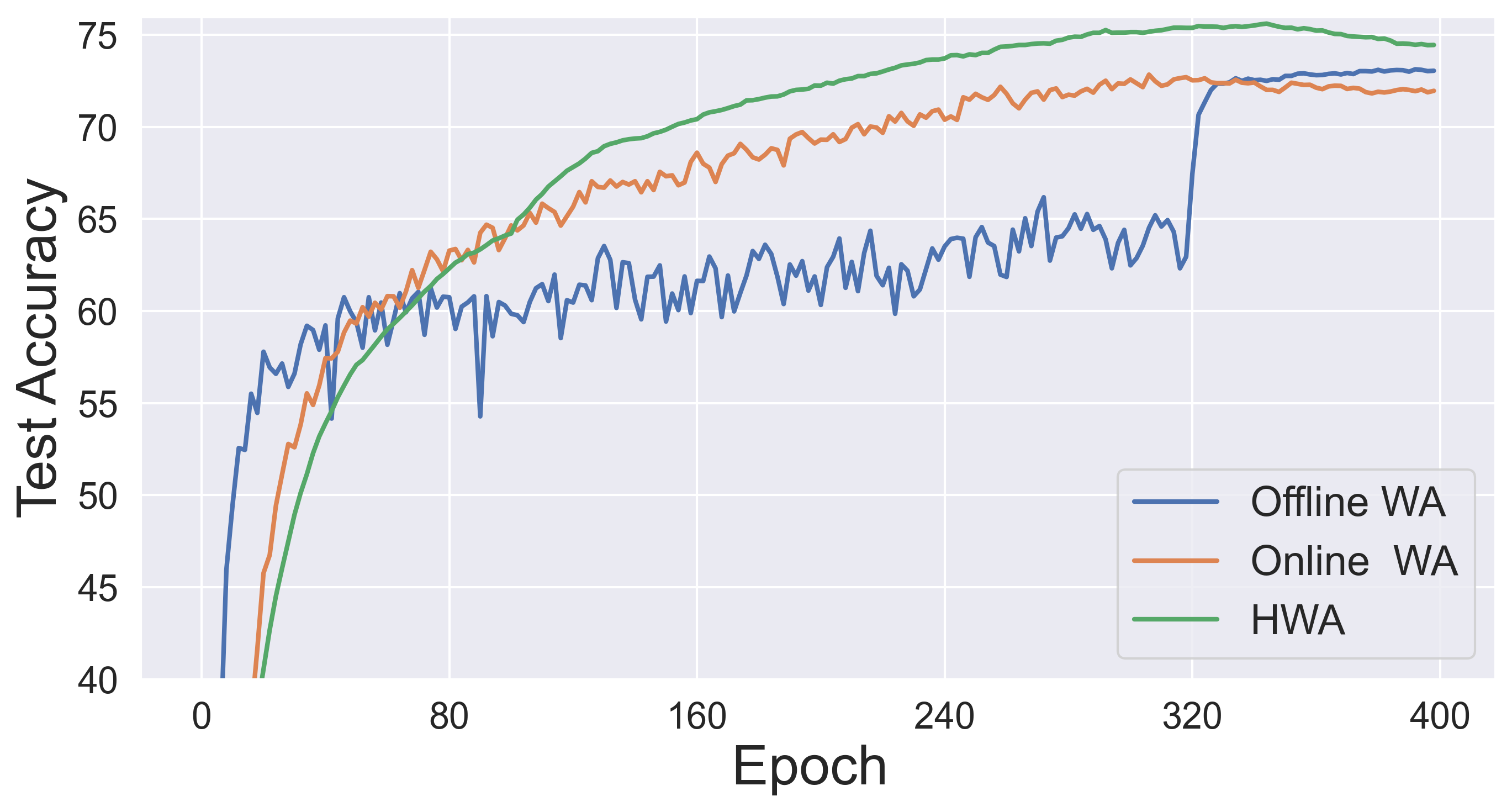}  
\caption{
The  top-1  test accuracy of online WA, offline WA, and HWA on CIFAR100 for  ResNet56.
}
\label{fig:VS2}
\end{figure}

Note that, \emph{ both online and offline WA is a special case of our proposed HWA}.  In Table~\ref{tab:tbcompare} we also provide a comparison of online WA, offline WA, and our proposed HWA. While online and offline WA can only serve a single purpose, \ie,  improving  training efficiency or  generalization ability,   the hybrid paradigm of HWA is  able to achieve both of them with better performance. 
As an example, we plot the   top-1  test accuracy of online WA, offline WA, and HWA  on CIFAR100  for ResNet56 as a function of training time in Figure~\ref{fig:VS2}, where online WA and HWA apply a cosine annealing learning rate throughout the training process, and offline WA starts averaging  from the $320_{th}$ epoch with a constant learning rate.

In sum, the contributions of this work are as follows.
\begin{enumerate}
    \item  We present a novel weight averaging scheme termed Hierarchical Weight Averaging (HWA)  which incorporates { two different types of WA techniques (\ie, online and offline WA), which are seldom associated with each other, into a unified training framework.}  By leveraging our  designed \emph{low-frequency online WA} and  \emph{slide-window based offline WA} in a hierarchical manner, HWA achieves several appealing benefits: 1) Unlike online or offline WA, which only serves  a single purpose, HWA can improve training efficiency and generalization ability at the same time. 2) HWA outperforms online WA and offline WA in terms of training efficiency and generalization ability, respectively.
    
    \item {We also empirically analyze the issues faced by existing offline WA (\eg, the generalization performance is sensitive to the learning rate scheduler (WA\_LR)) and  how our HWA address them  in Section~\ref{sec:discuss}.}
    
    \item Extensive experiments on  CIFAR~\cite{krizhevsky2009learning}, Tiny-ImageNet ~\cite{deng2009imagenet}, CUB200-2011~\cite{wah2011caltech},   CalTech256~\cite{griffin2007caltech} and ImagenNet~\cite{krizhevsky2012imagenet} have been conducted to demonstrate the effectiveness of the proposed HWA. According to the results, HWA consistently outperforms the other related  methods  for a variety of architectures.  { For example, it outperforms the best state-of-the-art competitor by $1.16\%$  on average  across multiple architectures on CIFAR100 dataset. More experimental results could be found in Section~\ref{sec:exp}.}
\end{enumerate}
The rest of the paper is organized as follows. In Section~\ref{sec:relatedwork}, we introduce the related works. Section~\ref{sec:hwa} gives the detailed procedure of our proposed HWA. In Section~\ref{sec:exp}, we discuss how HWA improves the training efficiency and the generalization ability of DNNs. In Section~\ref{sec:exp}, we evaluate HWA on several benchmark 
datasets and  compare it with other related methods. Finally, Section~\ref{sec:conclusion} concludes the paper.

\section{Related Work}
\label{sec:relatedwork}
\paragraph{Offline  Weight Averaging}
Averaging  a set  of weights traversed by SGD offline  can date back several decades in convex optimization~\cite{ruppert1988efficient,polyak1992acceleration}, but is not typically used to train neural networks. These techniques reduce the impact of noise  and improve  convergence  rates.  The Stochastic Weight Averaging (SWA)~\cite{izmailov2018averaging} is proposed  to exploit the flatness of training objectives and improve generalization specific to deep learning. Based on the observation that weights of the networks sampled by FGE~\cite{huang2017snapshot} are on the periphery of the most desirable solutions, SWA~\cite{izmailov2018averaging} creates a network with the averaged weights instead of forming an ensemble.  While SGD tends to converge to the boundary of the low-loss region, SWA is able to find a point centered in this region with better generalization.  

Following up SWA, various extensions  have been proposed for different purposes. SWAG~\cite{maddox2019simple}  can approximate Bayesian model averaging in Bayesian deep learning and achieves the state-of-the-art uncertainty calibration results in various settings. SWALP~\cite{yang2019swalp} can match the performance of full-precision SGD training with quantized parameters. SWAP~\cite{gupta2020stochastic}  greatly speeds up the training of neural networks by using large batch sizes. { With a different motivation, SWAD~\cite{cha2021swad} and DiWA~\cite{rame2022diverse} were proposed to improve the out-of-distribution (OOD) generalization performance of DNNs. In~\cite{wang2021memory}, the authors applies  WA in the network pruning task.} 

\paragraph{Online  Weight Averaging}  The classical parallel mini-batch  SGD~\cite{dekel2012optimal,li2014communication} utilizes distributed hardwares to achieve  fast and efficient training of large-scale DNNs. It typically samples  gradients of  different models  in parallel and updates each model’s weights using the averaged gradient in one step.   The parallel mini-batch SGD can be interpreted as  an online WA,  where  the local model  updates its parameters to the averaged weights at every iteration.

However, this algorithm imposes heavy communication overhead as it requires exchanging of local gradient information at every iteration. To reduce the communication overhead, model averaging techniques ~\cite{lin2018don,mcmahan2017communication,mcdonald2009efficient,zhang2016parallel,stich2018local,zinkevich2010parallelized} suggest performing the  weight synchronization operation every $H$ iterations. In order to balance the communication overhead and convergence speed, the value of parameter synchronization period $H$  considered in these works  is also relatively small, (\eg, $[1,64]$ in~\cite{lin2018don}, $10$ in~\cite{zhang2016parallel}, $[4,32]$ in ~\cite{yu2019parallel}).  One extreme case is the one-shot averaging~\cite{zinkevich2010parallelized}, where  weights are averaged only at the last iteration but  may yield inaccurate solutions for certain non-convex optimization~\cite{zhang2016parallel}.  Theoretical analysis of the convergence speed of  online WA could be found  in~\cite{zhou2017convergence,stich2018local,yu2019parallel}.  {While  existing offline/online WA methods  perform either offline parameter averaging or online parameter averaging, HWA performs both of them during the training procedure.}

\paragraph{Other Related Works}
The  Lookahead optimizer~\cite{zhang2019lookahead}
maintains a set of slow weights  and fast weights, and updates the slow weights with the fast weights every $H$ steps. The stochastic gradient descent with restarts (SGDR)~\cite{loshchilov2016sgdr} first introduces the idea of scheduling the learning rate cyclically as a modification to the common linear or step-wise decay strategy. Some studies~\cite{huang2017snapshot} point out that SGDR can escape from local minimum and explore other regions.
Recently, a training strategy named   Improved Training of Convolutional Filters (RePr)~\cite{prakash2019repr}  was proposed to improve the network on the filter level.  RePr first freezes the overlapped filters to train the sub-network,  and then restores the frozen filters and continues to train the full network.   With a similar motivation, Filter Grafting (FG)~\cite{meng2020filter} uses entropy-based criteria to measure the information of filters and grafts external information from other networks trained in parallel into invalid filters. { SAM~\cite{foret2020sharpness} is a novel optimizer that improves
model generalization by simultaneously minimizing loss value and loss sharpness. }

\section{Detailed Procedure of HWA}
\label{sec:hwa}
In this section, we  introduce the detailed procedure of HWA and  leave the analysis of how HWA works in Section~\ref{sec:discuss}.  The general idea of HWA is given in Figure~\ref{fig:L123}(c).  In Algorithm~\ref{alg:hwa1} and \ref{alg:hwa2}, we also provide the  pseudo-code  of the  online   and  offline  modules of HWA, respectively.

\subsection{Online Module of HWA} At the online stage,  HWA trains $K$ models in parallel  with different  sampling orders. A standard SGD optimizer $\mathcal A$ with a regular decaying learning rate scheduler  is used to evolve each model by performing $H$ sequential SGD updates to batches of  training examples $\psi$ sampled from the dataset $\mathcal D$:
\begin{align*}
\overbrace{\mathbf W^k_{e,t}}^{{\emph{Inner  Weights}}}&\leftarrow \mathbf W^k_{e,t-1}+\mathcal A\left(\mathcal L, \mathbf W^k_{e,t-1},\psi_{e,t}^k\right)\\
\mbox{where~}&k\in\{1,2,\ldots, K\}~,t\in \{1,2,\ldots, H\}
\end{align*}
Here $\mathcal L$ is the loss function, and $e$ denotes $e_{th}$  averaging or synchronization cycle.   After   each model is updated for $H$ iterations ($\mathbf W^k_{e,0}\rightarrow \mathbf W^k_{e,H}$), HWA takes the average of the weights of the $K$ models in parallel  to obtain the averaged weights for the $e_{th}$ synchronization cycle
$$
\overbrace{\overline{\mathbf W}_{e}}^{\emph{Outer  Weights}}=\frac{\sum_{k=1}^K\mathbf W_{e,H}^k }{K}
$$
Once completed, it  synchronizes  the weights of each model at the start of the next synchronization cycle $\mathbf W_{e+1,0}^k$ to the averaged weights
$$
\mathbf W_{e+1,0}^k\leftarrow \overline{\mathbf W}_{e}~\mbox{~where~}k\in\{1,2,\ldots, K\}
$$
Since   $\mathbf W_{e+1,t}^k$ is updated within a synchronization cycle and $\overline{\mathbf W}_{e}$ is calculated outside of a synchronization cycle, we name them \emph{Inner Weights} and \emph{Outer Weights}, respectively.   

\begin{algorithm}[h!]
    \caption{Online Module of HWA}
    \label{alg:hwa1}
    \KwIn{Synchronization Period $H$, Optimizer $\mathcal A$, Loss Function $\mathcal L$, Number of Models $K$,  Maximum Iterations  $T$}
    Initialize Inner Weight $\mathbf W_{0,0}^k$ for $k\in\{1,2,\ldots, K\}$\\
    \For{$i = 1$ to $T$}{
        $e\leftarrow \lfloor (i-1)/H\rfloor$  ($e_{th}$ synchronization cycle)\\
        $t\leftarrow  i-e\times H$ ($t_{th}$ step in the $e_{th}$ synchronization cycle)\\
     
        \For{k = 1 to $K$}{
            Sample a batch of data $\psi_{e,t}^{k}$ for model $k$ \\
            $\mathbf W_{e,t}^k\leftarrow \mathbf W_{e,t-1}^k+\mathcal A\left(\mathcal L, \mathbf W_{e,t-1}^k,\psi_{e,t}^k\right)$\\
        }
        \If{$t=H$}{
            $\overline{\mathbf W}_{e}=\frac{\sum_{k=1}^K\mathbf W_{e,t}^k }{K}$\\
            Save checkpoint $\overline{\mathbf W}_{e}$\\
          \For{k = 1 to $K$}{
               $\mathbf W_{e+1,0}^k\leftarrow \overline{\mathbf W}_{e}$\\
            }
          }
    }
\end{algorithm}




      



\subsection{Offline  Module of HWA}  At the offline stage,  HWA also further averages the outer weights $\overline{\mathbf W}_e$  saved at different synchronization  cycles during the online stage.  Unlike offline WA,  HWA does not require  a large  constant or cyclical learning rate at the end of the training process.   From our experiments,  we find it works well with a regular decaying learning rate scheduler over the entire budget, \eg, a cosine decay or linear decay. 

While offline WA averages all the model checkpoints sampled from the time it starts averaging, HWA uses a slide window with a length of $I$ and  takes the average of outer weights  at  different synchronization cycles within the window. 
\begin{align*}
\overbrace{\overline{\overline{\mathbf W}}_{e}}^{\emph{HWA Weights}}=\frac{\sum_{t=e-I+1}^{e}}{I} \overline{\mathbf W}_{t}
\end{align*}
In practice,   the offline WA module does not necessarily average all the outer weights $\overline{\mathbf W}_{e}$ within the slide window. For example, it can take the average of  outer weights $\overline{\mathbf W}_{e}$ with index in multiples of a particular integer, \eg, $\{\overline{\mathbf W}_{3},\overline{\mathbf W}_{6},\overline{\mathbf W}_{9},\ldots\}$.  Besides, when we have sufficient training budget,   we can try multiple  possible $I\in \mathbb I=\{I_1,I_2,\ldots\}$  as the length of slide window also has a significant impact on the performance of the final averaged model $\overline{\overline{\mathbf W}}_{e}$.
\begin{algorithm}[h!]
    \caption{Offline  Module of HWA}
    \label{alg:hwa2}
    \KwIn{ Checkpoints of Outer Weights $\overline{\mathbf W}_e$ for $e\in \{1,2,3...\}$, Slide Window Length $I$.}
    
    \For{$e = 1,2,\ldots$}{
   $\overline{\overline{\mathbf W}}_{e}=\frac{\sum_{t=e-I+1}^{e}}{I} \overline{\mathbf W}_{t}$\\
    Update batch normalization statistics if the DNN uses batch normalization
    }
\end{algorithm}

\section{How HWA Works?}
\label{sec:discuss}

In this section, we   discuss how  HWA  improves the training efficiency and  generalization ability of DNNs. 
\subsection{Convergence Condition of SGD} 
\label{sec:convsgd}

In order to study how HWA improves SGD, here  we first derive the convergence condition of SGD.  Let $R_N(\mathbf W_{e,t})=\frac{1}{N}\sum_{i=1}^N \mathcal L(\mathbf W_{e,t},\psi_i)$ denote the empirical risk of model  $\mathbf W_{e,t}$ and $\psi_i=(x_i,y_i)$ denote a single training sample.  Suppose the batch size is $B$, then the gradient of a batch of samples is  $$
g(\mathbf W_{e,t},{\psi}_{e,t})=\frac{1}{B}\sum_{i=1}^B \nabla_{W} \mathcal L(\mathbf W_{e,t},{\psi}_{i})\approx \nabla_W R_N(\mathbf W_{e,t})$$
On the assumption of Lipschitz-continuous gradient with Lipschitz constant $L$ and the update  at each iteration is $\Delta W=-\eta g(\mathbf W_{e,t},{\psi}_{e,t})$, we have 

\begin{align*}
&||\nabla_W R_N(\mathbf W_{e,t}+\Delta W)-\nabla_W R_N(\mathbf W_{e,t})||\leq L ||\Delta W||_2\\
\Rightarrow  &  R_N(\mathbf W_{e,t}+\Delta W)- R_N(\mathbf W_{e,t}) \\ & \leq  \frac{1}{2}L ||\Delta W||_2^2+\nabla_W R_N(\mathbf W_{e,t})^T \Delta W\\
\Rightarrow  &  R_N(\mathbf W_{e,t+1})- R_N(\mathbf W_{e,t})\\ &\leq \frac{1}{2}L \eta^2 || g(\mathbf W_{e,t},{\psi}_{e,t})||_2^2-\eta \nabla_W R_N(\mathbf W_{e,t})^T g(\mathbf W_{e,t},{\psi}_{e,t}),\\
\end{align*}

Suppose $\Psi_{e,t}$ denotes the variable of  mini-batch samples,  the empirical risk is expected to decrease  if 
\begin{align*}
 & \mathbb E_{\Psi_{e,t}}\left[ R_N(\mathbf W_{e,t+1})\right]-R_N(\mathbf W_{e,t}) \leq 0\\
 \Rightarrow & \frac{1}{2}\eta L \mathbb E_{\Psi_{e,t}}\left[||g(\mathbf W_{e,t},\Psi_{e,t})||_2^2\right] \\
 & - \nabla_W R_N(\mathbf W_{e,t})^T \mathbb E_{\Psi_{e,t}}\left[g(\mathbf W_{e,t},\Psi_{e,t})\right]\leq 0\\
  \Rightarrow &\frac{1}{2}\eta L ||\mathbb E_{\Psi_{e,t}} \left[g(\mathbf W_{e,t},\Psi_{e,t})\right]||^2_2 \\&+\frac{1}{2}\eta L \mathbb E_{\Psi_{e,t}}[||g(\mathbf W_{e,t},\Psi_{e,t})-\nabla_W R_N(\mathbf W_{e,t})||_2^2]&\\& \leq  \nabla_W R_N(\mathbf W_{e,t})^T \mathbb E_{\Psi_{e,t}}\left[g(\mathbf W_{e,t},\Psi_{e,t})\right]\\
  \Rightarrow & \frac{1}{2} L \mathbb E_{\Psi_{e,t}}[||g(\mathbf W_{e,t},\Psi_{e,t})-\nabla_W R_N(\mathbf W_{e,t})||_2^2]\\& \leq 
  (\frac{1}{\eta}-\frac{1}{2} L)||\nabla_W R_N(\mathbf W_{e,t})||^2_2
\end{align*}
For a fixed dataset $\mathcal D$ and model $\mathbf W_{e,t}$,  the right hand side  $(\frac{1}{\eta}-\frac{1}{2} L)||\nabla_W R_N(\mathbf W_{e,t})||^2$ is determined by the learning rate $\eta$,  and left hand side $ \frac{1}{2} L \mathbb E_{\Psi_{e,t}}[||g(\mathbf W_{e,t},\Psi_{e,t})-\nabla_W R_N(\mathbf W_{e,t})||_2^2]$ is controlled by the batch size.  As long as the learning rate $\eta$ stays constant,  model $\mathbf W_{e,t}$ would converge only if the mini-batch gradient $g(\mathbf W_{e,t},\Psi_{e,t})$ estimates $\nabla_W R_N(\mathbf W_{e,t})$ accurately enough. 
\begin{figure}[t!]
\centering
\includegraphics[width=0.45\textwidth]{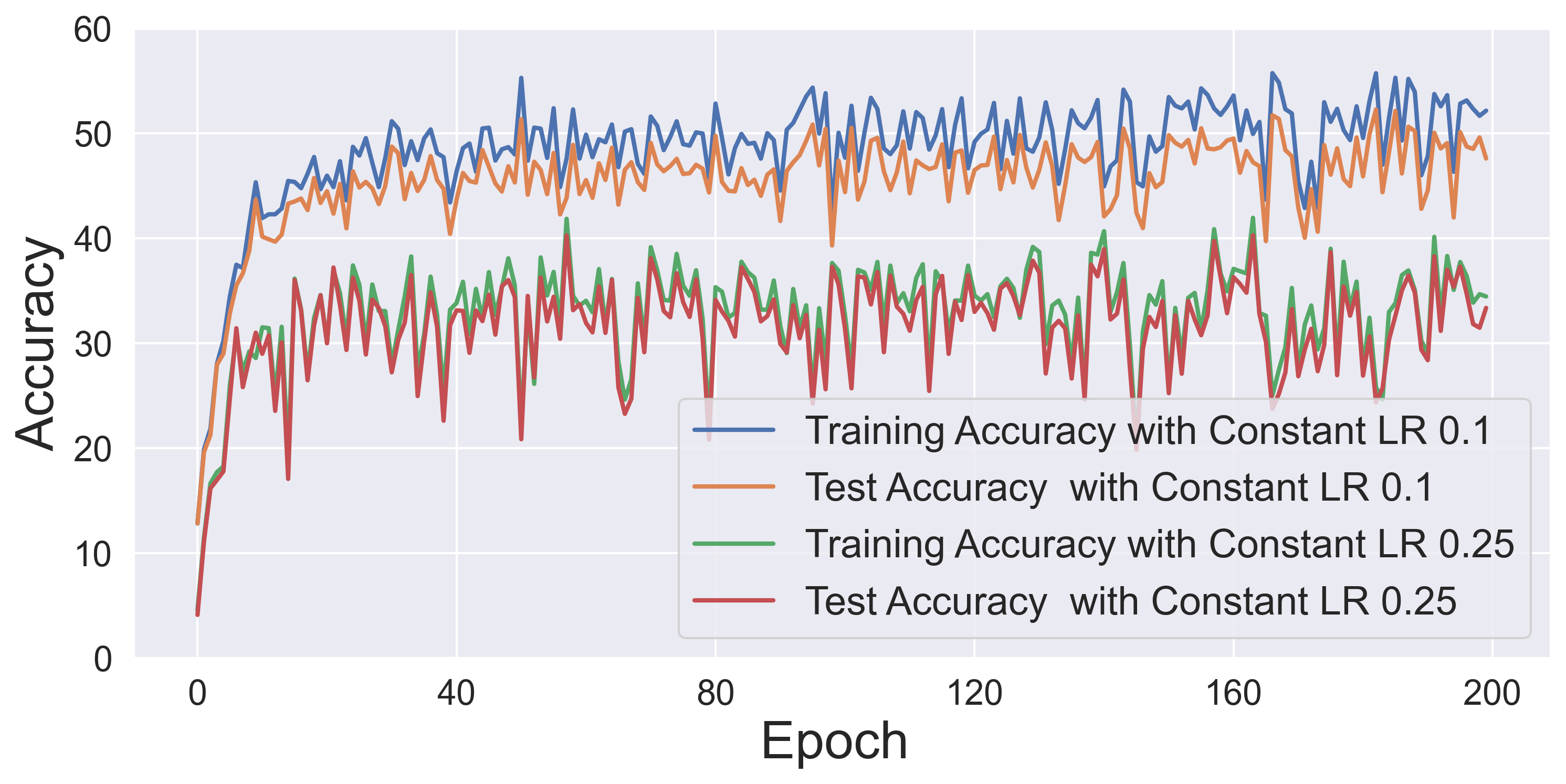}
\caption{The top-1 accuracy of ResNet20 on CIFAR100 with different constant learning rates.
}
\label{fig:constlr}
\end{figure}

Therefore, with a constant learning rate $\eta$, the optimizer would converge to a certain point and then start to oscillate and  explore around the minima.  This phenomenon is clearly shown in Figure~\ref{fig:constlr}, where we plot the training and test accuracy of ResNet20 on CIFAR100 when it is trained by SGD with a constant learning rate $\eta\in \{0.1,0.25\}$. As we can see, the training and test accuracy start to oscillate  after dozens of epochs, which indicates that the optimizer is exploring around the local minima instead of progressing  to the local minima.

\subsection{How HWA Improves Training Efficiency} 
\label{sec:discussefficiency}
Though  the loss functions of DNNs are known to be nonconvex~\cite{choromanska2015loss}, the loss surfaces  over the trajectory of SGD are approximately convex~\cite{goodfellow2014qualitatively}. According to the convergence condition,  SGD would oscillate and explore around the local minima after it converges to a certain point unless the learning rate $\eta$ further decays.  Therefore,  by taking the average of the weights of multiple model checkpoints, the averaged solution would  quickly progress to a point  closer to the minima with training lower loss.

\begin{figure}[t!]
        \centering
        \includegraphics[width=0.35\textwidth]{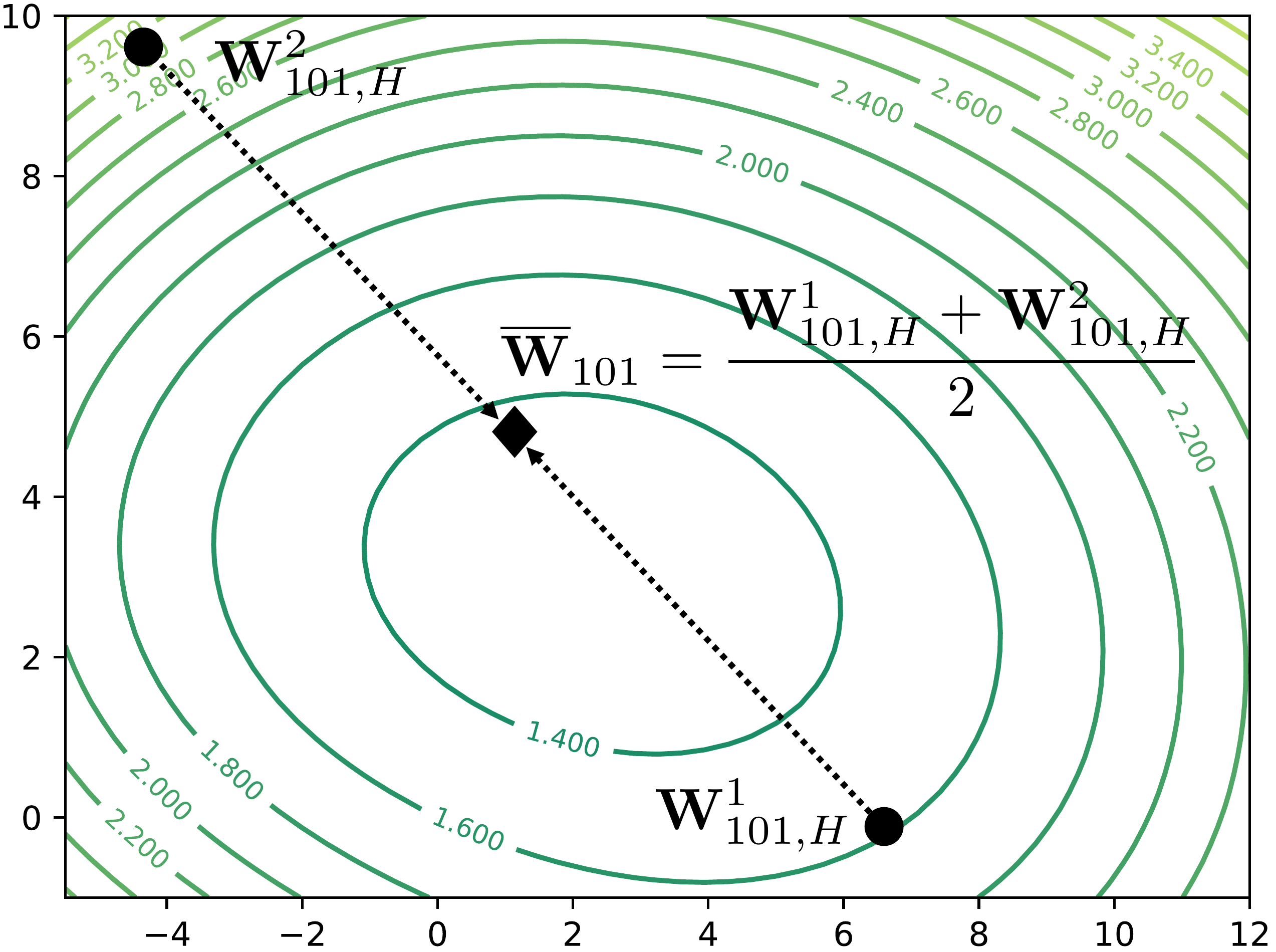}
        \caption{The projection of  the online averaging operation of HWA on the training loss surface of ResNet32 on CIFAR100.}
        \label{fig:onlinewavis}
\end{figure}

This scenario is shown in  Figure~\ref{fig:onlinewavis}, where we use the visualization tool~\cite{li2018visualizing}  to project  model checkpoint  $\mathbf W^1_{101,H}$ and $\mathbf W^2_{101,H}$ (two models in parallel)  and their corresponding averaged weights $\overline{\mathbf W}_{101}$ on the training loss surface of ResNet32 on CIFAR100. Note that,  $\mathbf W^1_{101,H}$ and  $\mathbf W^2_{101,H}$ are evolved by SGD for $H=391$ steps (one epoch) from the same starting point  $\overline{\mathbf W}_{100}$, and hence they are within the same local minima. Through the online averaging operation, the averaged solution $\overline{\mathbf W}_e$ immediately reaches  a location closer to the minima on the training loss surface.

\begin{figure}[t!]
        \centering
        \includegraphics[width=0.35\textwidth]{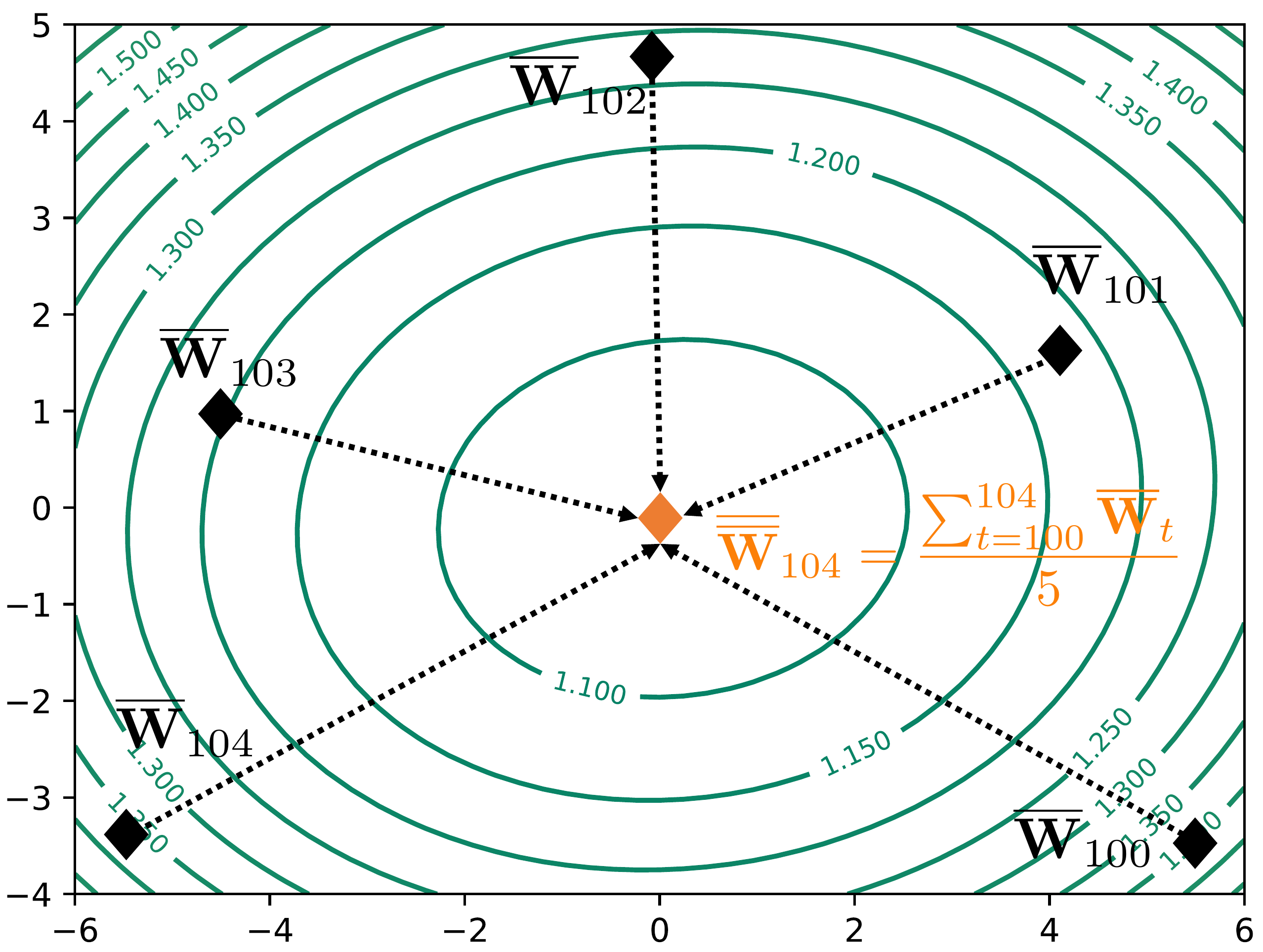}
        \caption{The projection of the offline averaging operation of HWA on the training loss surface of ResNet32 on CIFAR100.}
        \label{fig:offlinewavis}
\end{figure}

Similar to the online averaging operation of HWA,  the offline averaging operation of HWA also speeds up convergence in a similar way.  However as the offline averaging operation of HWA  is performed vertically at different synchronization cycles on top of  the outer weights $\overline{\mathbf W}_e$ obtained during the online stage,  the  HWA weights  $\overline{\overline{\mathbf W}}_e$  progresses to a point  closer to the minima than  $\overline{\mathbf W}_e$.  To  show  this scenario, we also use the visualization tool~\cite{li2018visualizing} to project the offline averaging operation of HWA in Figure~\ref{fig:offlinewavis}. 

\begin{figure}[h!]
\centering
\includegraphics[width=0.5\textwidth]{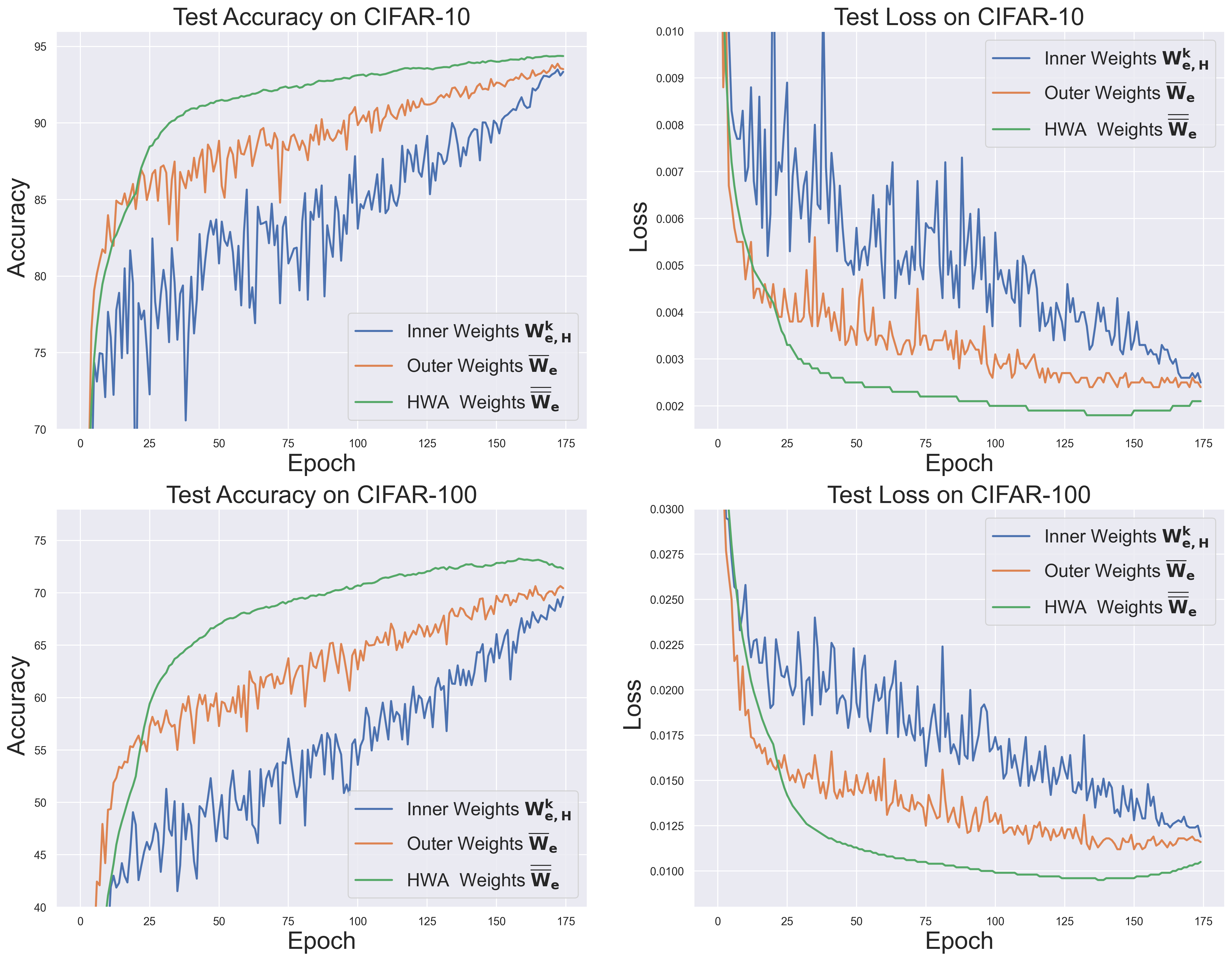}
\caption{Top-1 test accuracy and test loss of  \emph{Inner Weights} $\mathbf W_{e,H}^k$, \emph{Outer Weights} $\overline{\mathbf W}_e$, and  \emph{HWA Weights} $\overline{\overline{\mathbf W}}_e$ for ResNet32 on CIFAR10 and CIFAR100. The learning rate decays along a cosine curve.
}
\label{fig:mobv2}
\end{figure}

Therefore,  by performing the weight  averaging operations  in both  horizontal and vertical directions  hierarchically, HWA can progress to the local minima more quickly than online WA alone, which only performs the averaging operation horizontally. In Figure~\ref{fig:mobv2}, we plot the top-1 test accuracy and test loss of \emph{Inner Weights} $\mathbf W_{e,H}^k$, \emph{Outer Weights} $\overline{\mathbf W}_e$, and  \emph{HWA Weights} $\overline{\overline{\mathbf W}}_e$ for ResNet32 on CIFAR10 and CIFAR100, where the synchronization period is $H=391$ (\ie, one epoch) and  the length of slide window is $I=20$. As we can observe,  the \emph{HWA Weights}  $\overline{\overline{\mathbf W}}_e$  achieves faster convergence than \emph{Outer Weights} $\overline{\mathbf W}_e$ and inner weights  $\mathbf W_{e,H}^k$.

\subsection{How HWA Improves Generalization Ability}
\label{sec:discussgeneralization}
One popular hypothesis about  the \emph{\textbf{generalization ability}}\footnote{In this paper, we refer to  generalization ability as the ability  in predicting unseen, \emph{in-distribution} data (\eg, test data).}  of DNNs~\cite{cooper2018loss,draxler2018essentially,izmailov2018averaging} is that  the generalization performance of DNNs is correlated with the flatness of its local minima, and a sharp local minimum typically tends to generalize worse than the flat ones~\cite{keskar2016large}.  While most existing works (\eg,~\cite{izmailov2018averaging,zhang2019your}) attribute the better generalization ability of neural networks to a wider local minimum,  the location to which a network converges within  a local minimum also  greatly affects  the generalization performance.   As observed in~\cite{he2019asymmetric},  the loss landscape of  local minima is asymmetric and exists both sharp and flat directions, and the solution locates at the flat side has better generalization performance due to  the  random shift between the  training loss and test loss.

\begin{figure}[t!]
\centering
\includegraphics[width=0.4\textwidth]{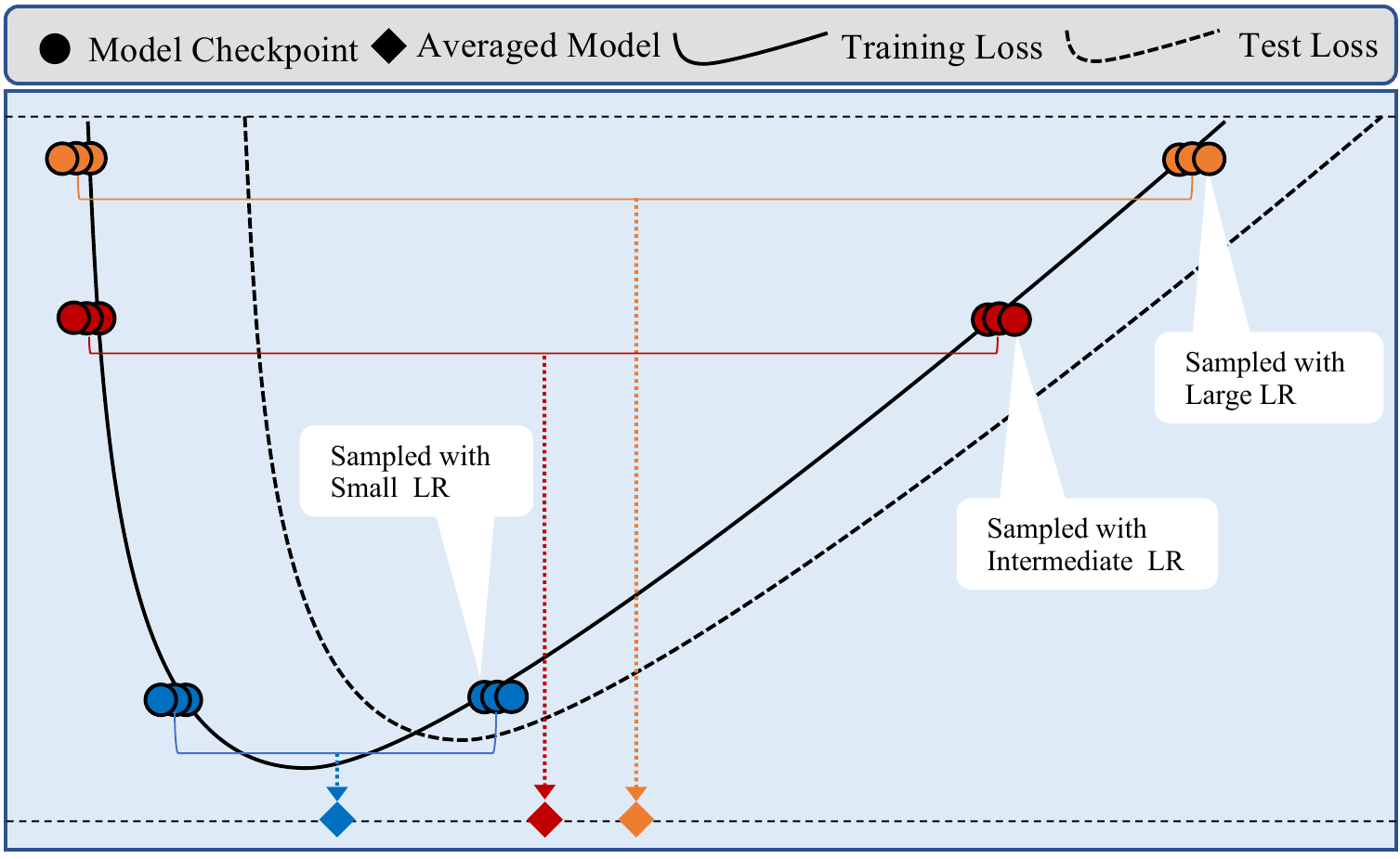}
\caption{A simple illustration of  why averaged model tends to be biased to the flat side of the local minima.
}
\label{fig:biasflat}
\end{figure}

\begin{figure}[h!]
\centering
\includegraphics[width=0.4\textwidth]{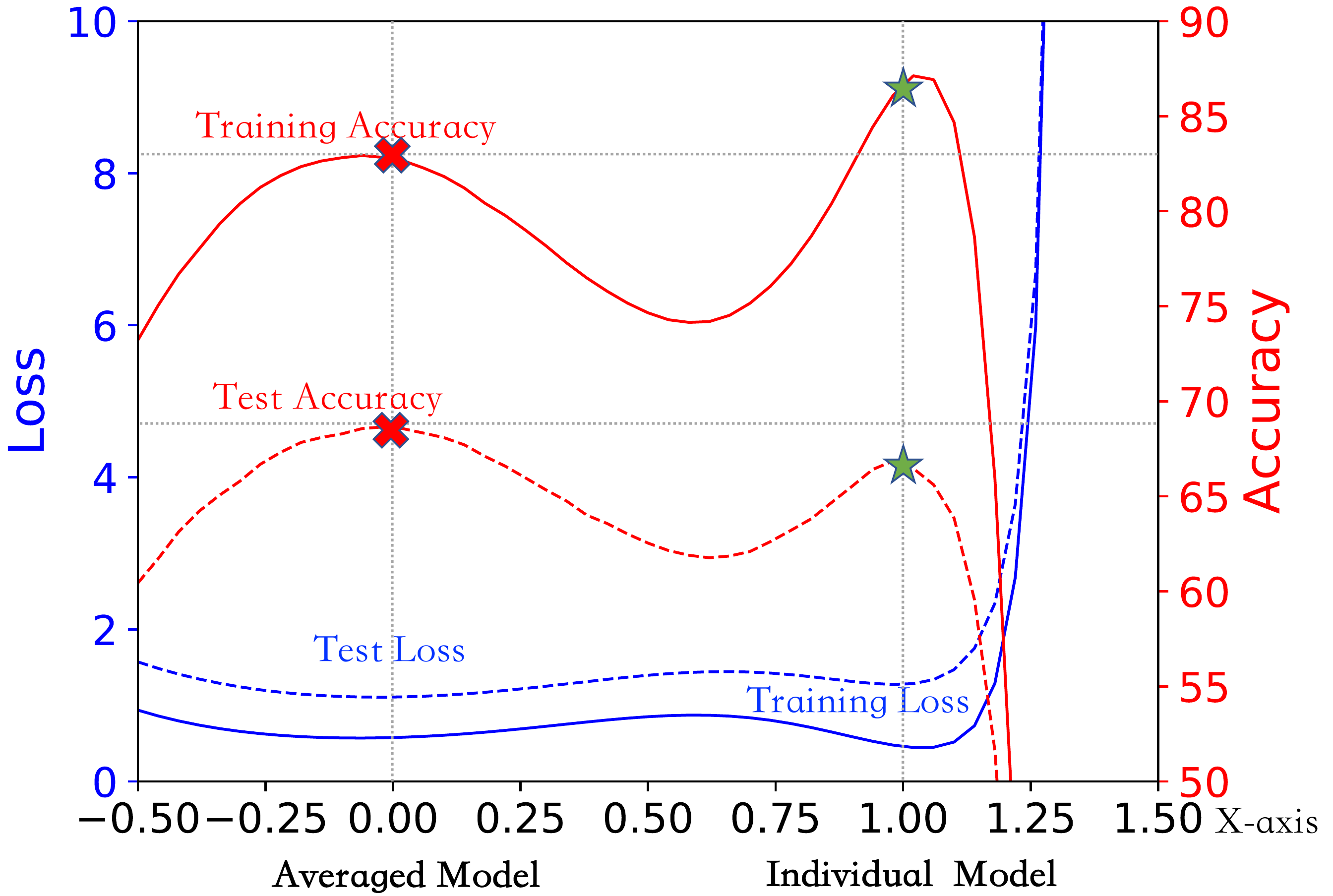}
\caption{The accuracy and loss   on CIFAR100 along the direction between an averaged model (location $x=0$) and an individual model (location $x=1$).
}
\label{fig:asymmetric}
\end{figure}

According to the convergence condition of SGD, when the model checkpoints are sampled with the same learning rate, they would be  distributed around  the local minima with similar empirical risks  (refer to Figure~\ref{fig:constlr}).  As a result, as illustrated in Figure~\ref{fig:biasflat},  the averaged model  tends to be  biased to the flat side of the local minima.  Due to the  shift between the training and test loss, the averaged model has better generalization ability than the model with the minimal training loss.  Such an example is shown in  Figure~\ref{fig:asymmetric}, which   plots  the accuracy and loss of ResNet20 on CIFAR100 along the direction between an averaged model   (location $x=0$) and an individual model   (location $x=1$).  Apparently,  all the models  between  $x=0$ and  $x=1$ are in the same local minima because there is no sharp drop in  training and test accuracy.  Despite the lower training accuracy,   the  averaged model  achieves higher test accuracy than the individual model, because it is biased to the  flat side of the loss landscape. This also implies that the training accuracy  is a poor indicator of the generalization ability.

Figure~\ref{fig:biasflat} also explains why offline WA is sensitive to the choice of learning rate used to sample model checkpoints. According to the convergence condition of SGD, the learning rate determines the empirical risk or loss of the sampled  model checkpoints (refer to Figure~\ref{fig:constlr}) and hence their locations on the loss landscape. Since the averaged model is obtained by taking the average of these model checkpoints, its  location within the local minima is also determined by the learning rate with which these model checkpoints are sampled.   As a result, a choice of learning rate that is either too small or too large  can be detrimental to  the  performance of offline WA, which leverages  a large constant or cyclical learning rate to sample model checkpoints. For example  in   Figure~\ref{fig:biasflat}, the averaged model obtained with an intermediate learning rate achieves the best test performance.

\begin{figure}[t!]
\centering
\includegraphics[width=0.35\textwidth]{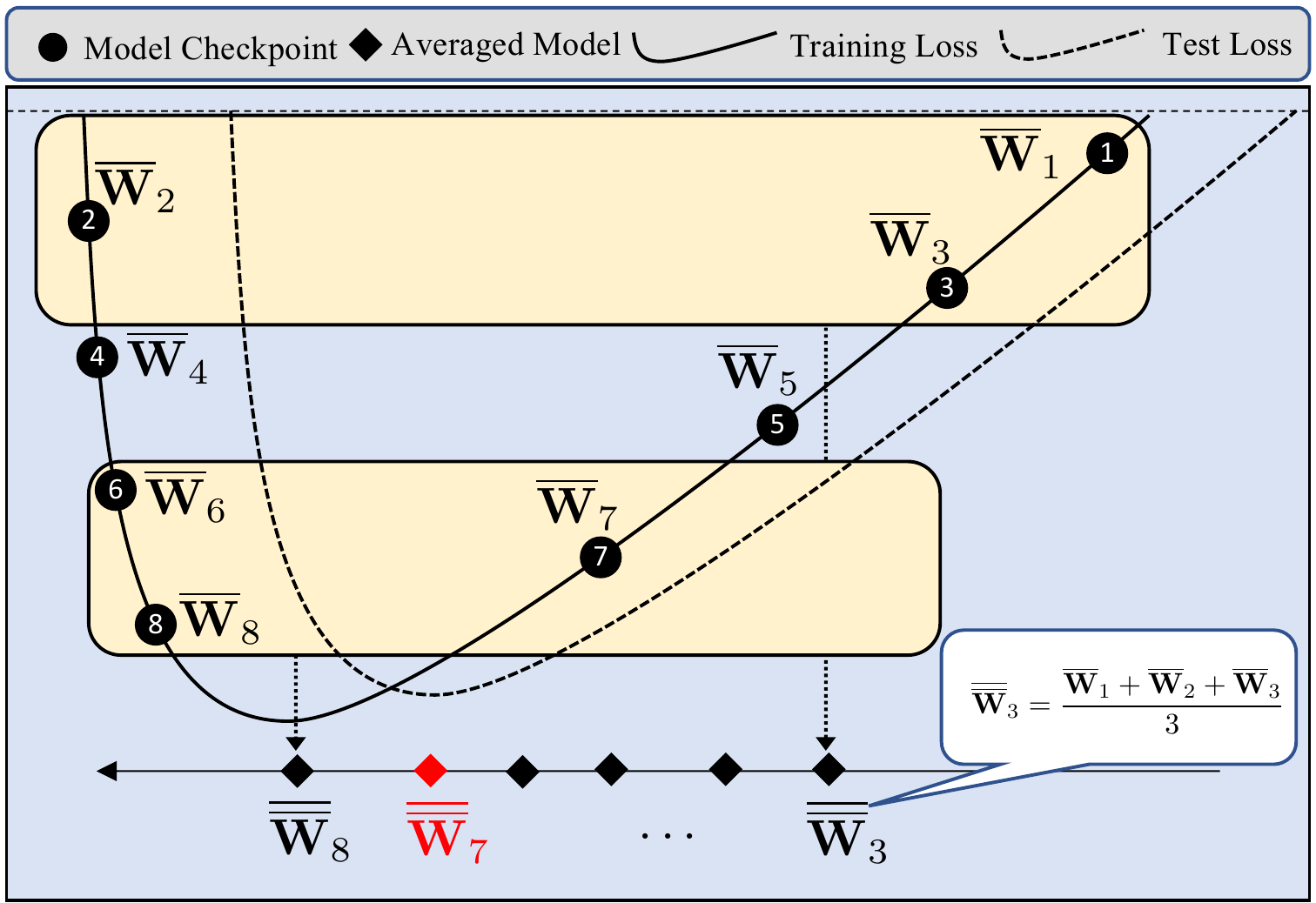}
\caption{The averaged model $\overline{\overline{\mathbf W}}_e$ at different  synchronization cycles is located at different positions on the loss landscape  
}
\label{fig:biasflat2}
\end{figure}

Unlike offline WA, as illustrated in Figure~\ref{fig:biasflat2}, HWA  takes the average of  model checkpoints  sampled with different learning rates  with a slide window of length $I$.  As a result, the averaged model $\overline{\overline{\mathbf W}}_e$ at different synchronization cycles is located at different positions  with different distances to the minima of the training loss surface.  Compared to offline WA,  HWA is able to explore more locations on the flat side of the local minima and hence is more likely  to find a better averaged model.  

\begin{figure}[h!]
\centering
\includegraphics[width=0.4\textwidth]{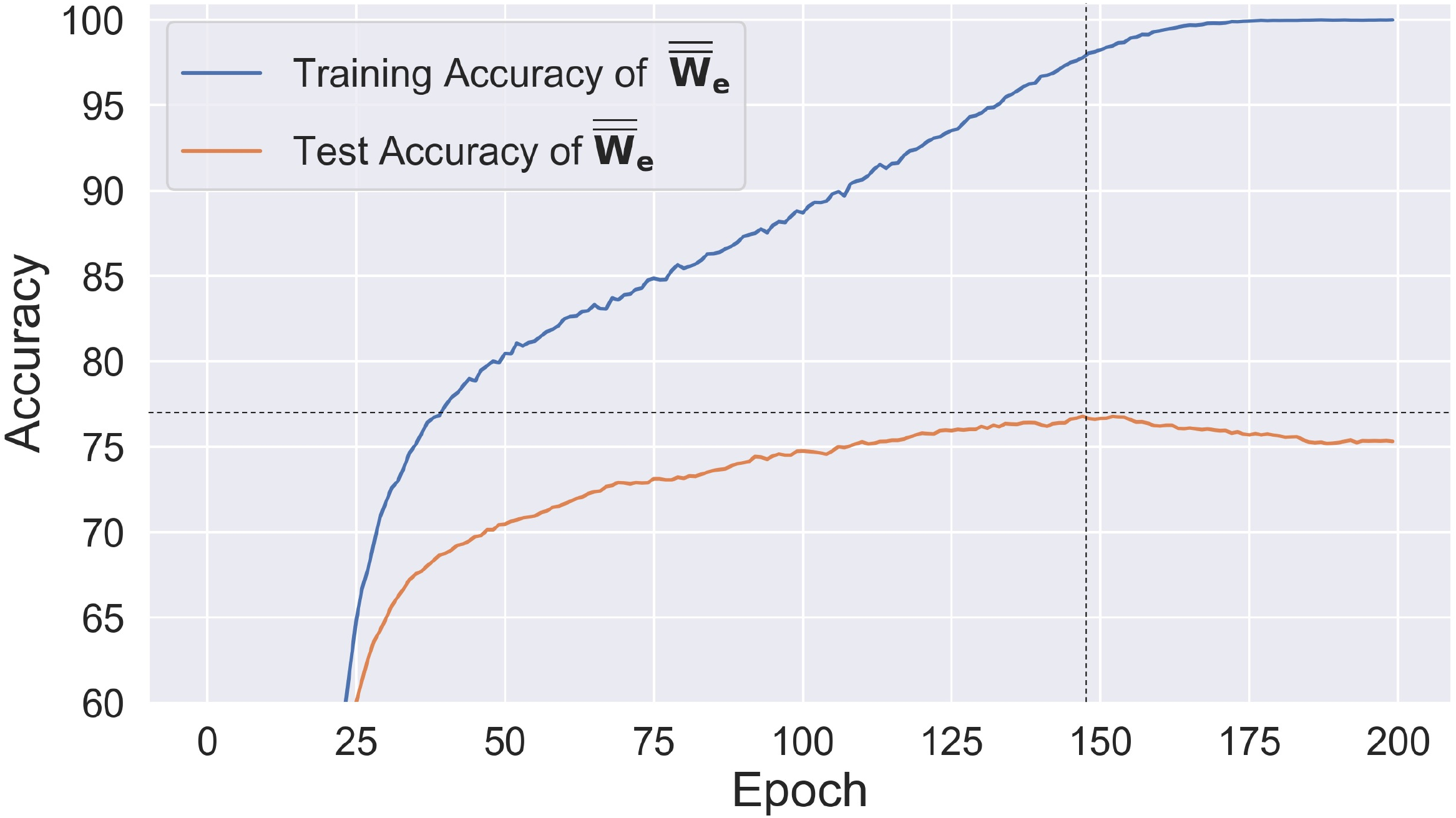}
\caption{The training and test accuracy of $\overline{\overline{\mathbf W}}_e$ as a function of epoch (the synchronization period is one epoch) for ResNet110 on CIFAR100.
}
\label{fig:biasflat3}
\end{figure}

Note that, the averaged model at the last synchronization cycle does not necessarily be the best averaged model. For example, as illustrated in Figure~\ref{fig:biasflat2}, the best averaged model  $\overline{\overline{\mathbf W}}_7$ is obtained at the $7_{th}$ synchronization cycle.  We also plot the training and test  accuracy of $\overline{\overline{\mathbf W}}_e$ as a function of epochs  for ResNet110 on CIFAR100 in Figure~\ref{fig:biasflat3}, where the synchronization period is one epoch and the length of slide window $I=20$. As we can observe, the  averaged model   $\overline{\overline{\mathbf W}}_e$  around the $150_{th}$ synchronization cycle achieves the best test accuracy.  Therefore, in practice, we  use techniques such as early stopping  to obtain the best averaged model.

Though the averaged model with the minimal training loss   does   not necessarily be  the averaged model with the best validation performance, in most cases,  the best averaged model is  located close to the minima of the training loss.  Therefore, it is important for the optimizer to explore more locations close to the  minima of the training loss.  With the help of the online WA module,  HWA progresses to  the regions
close to the minima of training loss more quickly than naive offline WA, which as a result,  helps it to find  a better averaged model.

\begin{figure}[h!]
\centering
\includegraphics[width=0.35\textwidth]{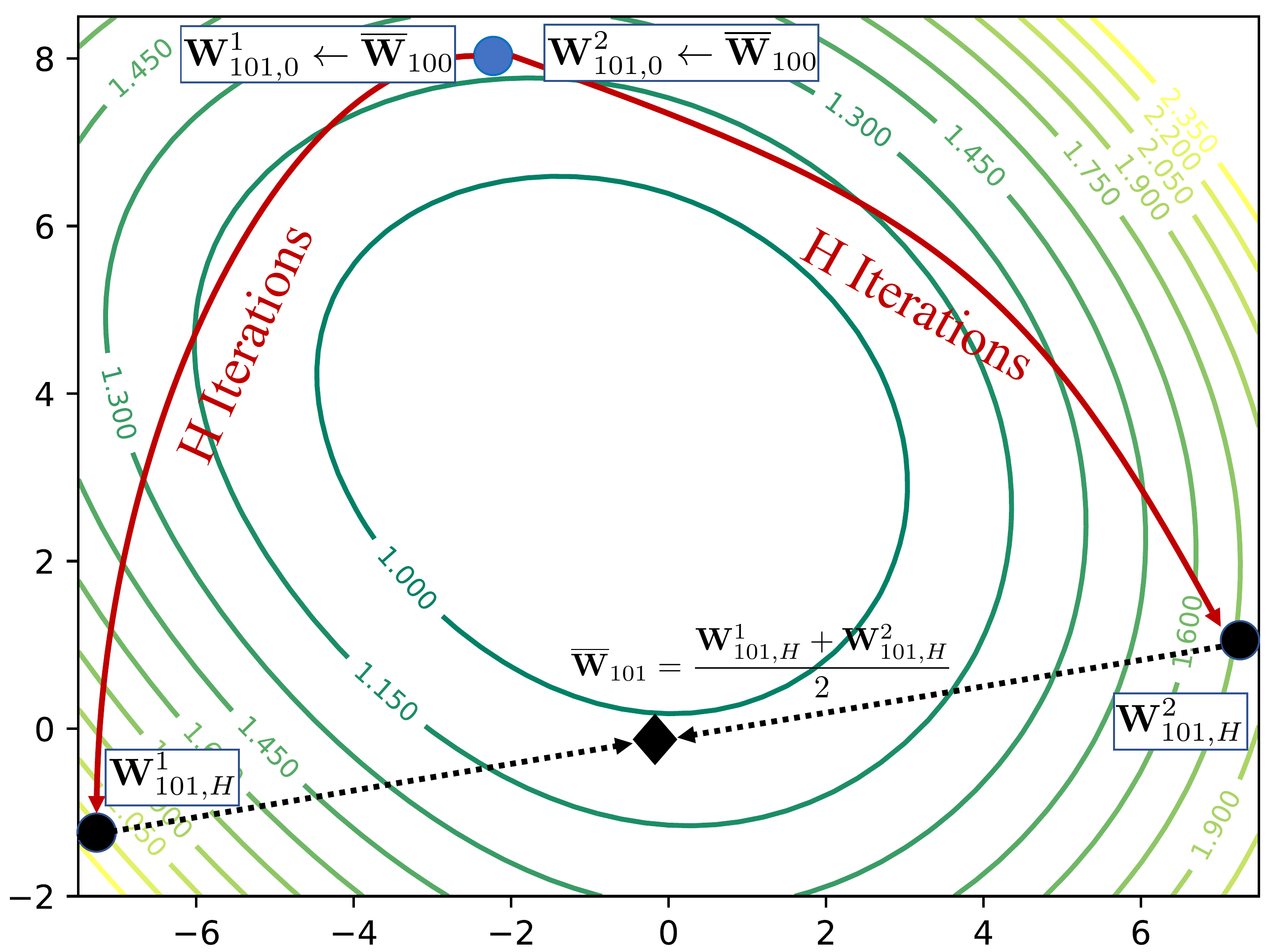}
\caption{The projection  of the  `restart' effect of HWA  on the training loss surface of ResNet32 on CIFAR100
}
\label{fig:restart}
\end{figure}

\noindent \textbf{Discussion:}  The   ‘restart’ mechanism~\cite{loshchilov2016sgdr,smith2017cyclical},  which switches the status of a  model to a lower fitting level to the training data (\ie, higher training loss),  is claimed to achieve a longer-term beneficial effect on generalization performance at the cost of short term benefit. While existing techniques~\cite{loshchilov2016sgdr,smith2017cyclical} restart the training by increasing the learning rate, we observe that the online averaging module of HWA  also achieves a similar  `restart' effect. We observe  that, in general, the loss of the averaged model $\overline{\overline{\mathbf W}}_e$ decreases over training time.  However,   within a synchronization cycle,  the loss of  inner weights $\mathbf W_{e,t}^k$ would surprisingly increase with iteration times $t$.  For example, as shown in Figure~\ref{fig:restart}, after  each model is updated by the optimizer for $H$ steps ($\mathbf W_{100,0}^k\rightarrow \mathbf W_{100,H}^k $), the training loss increases instead of decreasing.

Note that  the online averaging operation of HWA would result in a significant decrease in the empirical risk, \ie, 
\begin{align*}
R_N(\overline{\mathbf W}_e)&\ll R_N(\mathbf W_{e,H}^k),~
 \mathbf W_{e+1,0}^k \leftarrow \overline{\mathbf W}_e\\
 \Rightarrow &R_N(\mathbf W_{e+1,0}^k)\ll R_N(\mathbf W_{e,H}^k)
\end{align*}
As a result, the convergence condition  at the start of  the $e+1_{th}$ synchronization cycle does not hold any more for $\mathbf W_{e+1,0}^k$, and the optimizer $\mathcal A$ would receive more noise information than gradient information within the synchronization cycle. That's why the empirical risk $R_N(\mathbf W_{e+1,t}^k)$ increases in general with the iteration  times $t$ within a synchronization cycle. 

While there  is no consensus on  how the `restart' mechanism improves the training of DNNs,   we provide a possible explanation to the success of the  `restart' mechanism:  A neural network tends to generalize  easy training samples  and   memorizes the  ``hard'' samples through the massive parameters.   Once the  hard samples are memorized, their losses would  decrease to zero quickly, and hence the network  overfits the hard samples.  On the other hand, restart techniques switch the network status back  to a lower fitting level,  and hence the hard samples memorized by the network  are more likely  to be forgotten.

\section{Experiment}
\label{sec:exp}
~\begin{table*}[t!]
\normalsize
  \centering
  \begin{tabular}{l|c|cccccc}
     \toprule[2pt]
     Method       &Dataset       & ResNet20        &  ResNet32  &  ResNet56&  ResNet110 &  VGG16 & MobileNetV2\\
     \toprule[1pt]
    Baseline    &\multirow{6}{*}{CIFAR10}   	 & 91.60           &  92.60     & 	93.50	&	93.81	 &  93.25 & 93.89	 \\
	 CA        &~ & 92.51           &  93.09     &  93.60   &   93.91    &  94.01 & 94.11      \\

	 SWA   		  &~ & 92.67    &  93.38	 &	93.75   &	94.17    & 	94.18 &  94.80   \\

	 RePr~\cite{prakash2019repr} &~ &   93.10&  {\color{black}93.90}&-& 94.60&-&-  \\
	 FG~\cite{meng2020filter}&~&92.98&93.94& 94.73&94.96&94.26&95.03 \\
SAM~\cite{foret2020sharpness}&~&93.09&94.06&94.78&95.01&94.62&95.35\\
 HWA&~ &\textbf{93.46} &\textbf{94.41}  & \textbf{94.96}&  \textbf{95.19}  & \textbf{94.88}  & \textbf{95.46}\\
	 \midrule[1.5pt]
	 Baseline         &\multirow{6}{*}{CIFAR100}&	67.40	 &  69.82	  &	 71.55	  &  73.21	  &  72.55   &	73.39	 \\

	 CA            &~& 68.53    &  70.71     &  72.03    &  72.98    &  74.15   &  74.30      \\

	 SWA              &~& 69.03   &  70.20        &73.02	  & 71.57 &  73.92   &  76.90      \\
	 RePr~\cite{prakash2019repr}&~ &  68.90& 69.90& -& 73.60&-&-     \\
	 FG~\cite{meng2020filter}&~&68.84&71.28&72.83&75.27 &74.63&76.30 \\
 SAM~\cite{foret2020sharpness}&~&69.75&72.69&74.06&75.17&75.59&77.37\\
	 HWA   &~ & \textbf{70.79}  & \textbf{73.15} & \textbf{75.61} &\textbf{76.68}  &\textbf{77.12}  & \textbf{78.31}\\
	 \bottomrule[2pt]
  \end{tabular}
  \vspace{3pt}
  \caption{Comparison of HWA with other related methods on different networks. We use \textbf{bold} font to denote the method with the best result.
}
  \label{tab:cifarcomp}
\end{table*}
In this section, we evaluate our HWA on several benchmark datasets CIFAR10 and CIFAR100~\cite{krizhevsky2009learning}, Tiny-ImageNet~\cite{yao2015tiny}, CUB200-2011~\cite{wah2011caltech}, CalTech256~\cite{griffin2007caltech} and  ImageNet~\cite{krizhevsky2012imagenet},  and also compare it with other related methods. 

The two CIFAR datasets consist of images with a resolution of $32 \times 32$. CIFAR10 contains 10 classes and CIFAR100 contains 100 classes. Both datasets contain a train set of $50000$ images and a test set of $10000$ images. 
Tiny-ImageNet~\cite{yao2015tiny} comprises  $50000$  colored images over  $200$ classes with a resolution of $64 \times 64$. 
The  CUB-200-2011 contains $11788$ images of $200$ categories belonging to birds.  
The CalTech256  is an object recognition dataset comprising  $30607$ real-world images with different sizes over 256 object classes and an additional clutter class.
Imagenet is an image classification dataset with over  1.3 million images and contains 1000 classes.

In particular, the following methods are involved in the experiments. 
\begin{enumerate}
    \item \textbf{Baseline}:  the conventional SGD with step-wise learning rate decay ($0.1$ at every $60$ epochs).
    \item  \textbf{CA}: the cosine learning rate  strategy, which is claimed to outperform other learning rate schedulers.
    \item \textbf{FG}: filter grafting for deep neural networks~\cite{meng2020filter}. 
    \item  \textbf{RePr}: improved training of convolutional filters~\cite{prakash2019repr}.  
    \item \textbf{SWA}:  stochastic weight averaging (\ie, offline WA). 
     \item {\textbf{SAM}: shareness-aware minimization~\cite{foret2020sharpness}.}
    \item  \textbf{HWA}: our proposed HWA algorithm with   default synchronization period $H=N/B$, where $N$ is the number of training samples  and $B$ is the batch size. 
\end{enumerate}
\emph{Among the above mentioned techniques,  SWA is the  primary work we aim to compare because both  HWA  and SWA are  weight averaging strategies.}

\subsection{Comparison with Other Methods}
\label{sec:FOA}
In this section, we experimentally compare our HWA with other methods on  CIFAR10 and CIFAR100 datasets in Table~\ref{tab:cifarcomp}\footnote{The experimental results of  RePr~\cite{prakash2019repr} reported in Table~\ref{tab:cifarcomp} come from the corresponding paper because the code of  RePr~\cite{prakash2019repr} is not publicly available. ‘–’ denotes the result is not reported in the corresponding paper. 
For SWA~\cite{izmailov2018averaging} and FG~\cite{FGgit}, the experiments are conducted in our implementation using the code provided by the original group.}. We use bold font to highlight the best accuracy and blue color to denote the second-best result.  
For CIFAR, we evaluate our HWA on three popular networks: VGG, ResNet and MobileNetV2. All the networks are trained by SGD with a momentum of 0.9 and a weight decay of $5 \times 10^{-4}$.  
A  cosine  learning rate strategy  over the whole budget is used by default. The initial learning rate is set to 0.1 for ResNets and VGG, and 0.05 for MoibleNetV2.

The results in Table~\ref{tab:cifarcomp} demonstrate that HWA achieves the best performance among all the training schemes.  On average, it improves the baseline by around $1.8\%$ on CIFAR10 and $3.8\%$ on CIFAR100. In particular, it outperforms the  method with the second-best accuracy by $1.16\%$ on CIFAR100 on average. Compared to SWA, the performance of HWA is more stable. For example,  the test accuracy of SWA is even lower than the baseline  on CIFAR100 for ResNet110. On the other hand, with the same experiment setting, SWA achieves the second-best results for ResNet20, ResNet56 and MobileNetV2.  These results also support our claim that SWA is sensitive to the learning rate scheduler at the sampling stage.

~\begin{table}[h!]
  \centering
  \begin{tabular}{l|c|ccc}
     \toprule[2pt]
   Dataset        &     Method   & ResNet20        &  ResNet32& ResNet56\\
        \toprule[1pt]

      \multirow{4}{*}{CIFAR10} &Baseline     	 &  91.60          & 92.60    & 93.60		 \\
      
      ~&+CA     	 & 92.51           & 93.09     & 	93.60	 \\
       &+online  module    	 & 92.98           & 93.98     &  94.49		 \\
         &+offline module     	 & 93.46         & 94.41     & 	94.96	 \\
       
    \toprule[1pt]
      ~&Method     & ResNet110       &  VGG16&  MobileNetV2\\
        \toprule[1pt]
     \multirow{4}{*}{CIFAR10} 
     &Baseline     	 &  93.81          &   93.25  & 	93.89	 \\
     ~&+CA     	 & 93.91           & 94.01     & 	94.11	 \\
        &+online module     	 & 94.55           &   94.57   & 95.22		 \\
         &+offline module     	 &95.19          & 94.88     & 	95.46	 \\
	 \midrule[1.5pt]

	   &Method& ResNet20        &  ResNet32& ResNet56\\
    \toprule[1pt]
	  \multirow{4}{*}{CIFAR100}
	  &Baseline     	 &  67.40          & 69.82    & 71.55		 \\
	  ~&+ CA      &	68.53	 &  70.71	  &	 72.03	 	    \\
	   &+online module     	 & 69.42           &  70.91    & 	73.30	 \\
         &+offline module     	 & 70.79         &  73.15    & 	75.61	 \\
\toprule[1pt]
    ~  &Method     & ResNet110       &  VGG16&  MobileNetV2\\            
    
    \toprule[1pt]

      \multirow{4}{*}{CIFAR100} &Baseline     	 &  73.21          &  72.55   & 	73.39	 \\
      ~&+CA     	 & 72.98         & 74.15     & 	74.30	 \\
        &+online module     	 & 75.37          & 75.92     & 	76.28	 \\
         &+offline module     	 &  76.68        &  77.12    & 	78.31	 \\
	 \bottomrule[2pt]
  \end{tabular}
  \vspace{3pt}
  \caption{ Accuracy improvement  from  the online and offline averaging module of HWA on CIFAR10 and CIFAR100.  }
  \label{tab:onoff}
\end{table}
\begin{figure*}[t!]
        \centering
        \includegraphics[width=1\textwidth]{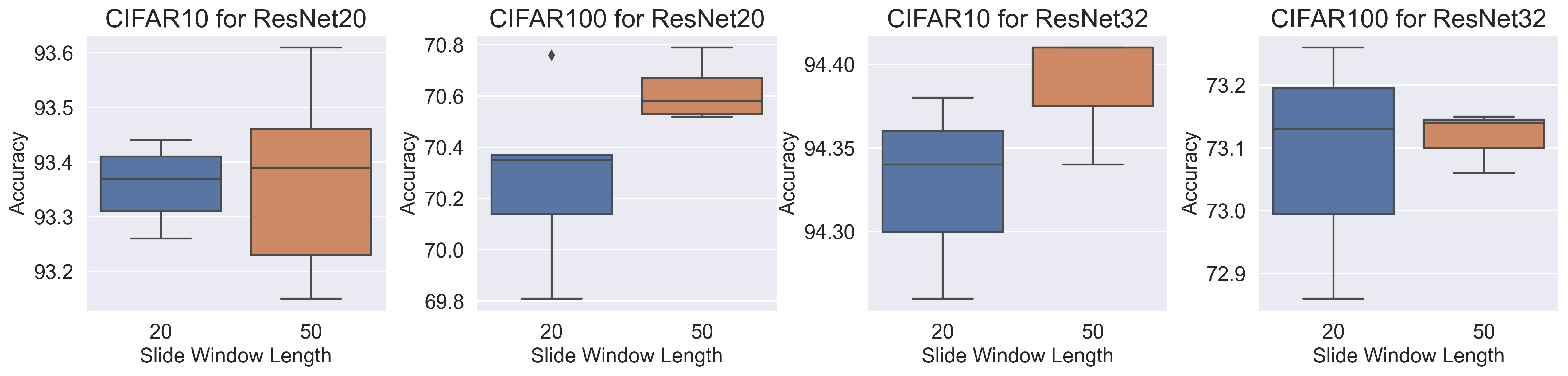}
        \caption{The box-plot of the test accuracy of ResNet20 and ResNet32 on CIFAR10 and CIFAR100. }
        \label{fig:variance}
\end{figure*}

\subsection{Contribution of Online and Offline WA Module }

Both the online  and offline WA module of HWA can improve the generalization performance of DNNs. In Table~\ref{tab:onoff}, we report the contributions of the online WA module of HWA  and the additional improvement from the offline WA module. On average, the online WA module  provides an  improvement around $0.8\%$ and $ 1.4\%$ on CIFAR10 and CIFAR100, respectively. By applying the offline WA module,   we obtain an  additional improvement around  $0.4\%$  and $1.7\%$ on CIFAR10 and CIFAR100, respectively. 

As we can observe, the performance of  the  online  WA module  is not stable when we apply it alone.  For example,  it provides almost no notable improvement for ResNet32 on CIFAR100.  However, by additionally  applying the offline WA module,  HWA outperforms CA by around $2.5\%$.
On the other hand,  the online WA module of HWA  improves CA by about $1.1\%$ for MobileNetV2 on CIFAR10. However, the additional gain in test accuracy  from the offline WA module is negligible (around $0.2\%$). In sum, the hybrid of  online and offline WA can improve the stability of HWA.

\subsection{Number of Models in Parallel}

In this section, we  explore how  the number of models in parallel  affects the final generalization performance of HWA.  In Table~\ref{tab:7}, we report the test accuracy of HWA on CIFAR10 and CIFAR100 with varying numbers of models  trained in parallel during the online stage. As we can observe, though the best test accuracy is achieved when   the number of models in parallel is  $K=3$ or $K=4$, the improvement in test accuracy is not very notable as $K$ increases from $2$ to $4$. Thus, in practice, we can simply  set $K=2$ when there is a limited training budget.

\begin{table}[ht!]
\small

  \centering
  \begin{tabular}{l|c|ccc}
  \toprule[2pt]
  Data &\diagbox{Model}{Number}& 2& 3& 4\\
  \toprule[1pt]
  \multirow{2}{*}{CIFAR100} & ResNet20 & 70.79 & \textbf{71.13} & 70.58 \\ 
                             ~& ResNet32 & 73.15    & 72.93 & \textbf{73.19} \\ 
  \midrule[1pt]
  \multirow{2}{*}{CIFAR10} & ResNet20 & 93.46  & \textbf{93.49} & 93.43 \\ 
                            ~& ResNet32 & 94.41 & 94.37 & \textbf{94.53} \\ 
  \bottomrule[2pt]
  \end{tabular}
  \vspace{3pt}
  \caption{Top-1 Accuracy with varying number of models  in parallel during the online stage.}
  \label{tab:7}
\end{table}

\subsection{Slide Window Length} 
In this section, we conduct several ablation experiments on CIFAR10 and CIFAR100 with ResNet20 and ResNet32 to study the influence from  the length of slide window. Each experiment is repeated  $5$ times and a slide window with length of $I=20$ or $I=50$ is considered in the  experiment. In Figure~\ref{fig:variance}, we present the box-plot of the experimental results. As we can observe, the length of the slide window  would exert different influences  on the test accuracy for different datasets. Therefore, we may achieve further improvement by  choosing a  suitable slide window length for different datasets and architectures for the offline WA module of HWA. 

\begin{figure*}[th!]
        \centering
        \includegraphics[width=1.0\textwidth]{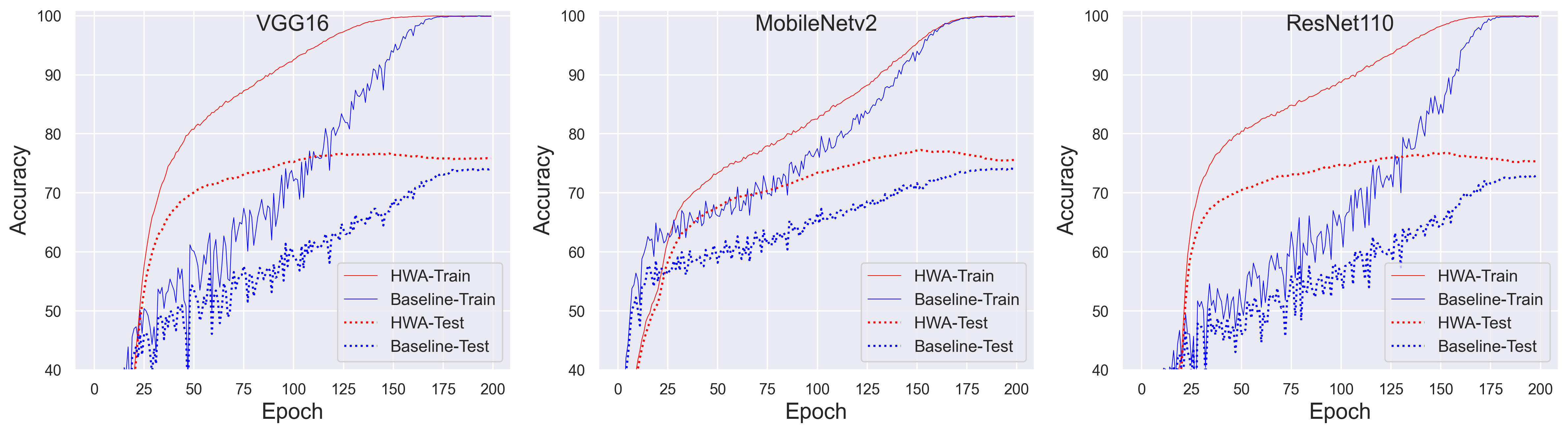}
        \caption{Training and test accuracy of HWA as a function of epochs  for VGG16, MobileNetV2 and ResNet110 on CIFAR100.}
        \label{fig:c100converge}
\end{figure*}

\subsection{Results on Other Datasets}
\label{sec:otherresult}
In this section, we evaluate our proposed HWA on several other datasets:  Tiny-ImageNet, CUB200-2011~\cite{wah2011caltech}, CalTech256~\cite{griffin2007caltech} and  ImageNet~\cite{krizhevsky2012imagenet}.  We use the models implemented in  Torchvision in the experiment. For comparison, we also report the  experimental results of  the baseline and SWA, which is the primary technique we aim to compare. Unless mentioned otherwise,    the experiment setting for CIFAR is also used in these experiments.

\noindent\textbf{Tiny-ImageNet:} In Table~\ref{tab:tinyimagenet} we show the top-1 test  accuracy  on Tiny-ImageNet, which comprises  $50000$  colored images over  $200$ classes with a resolution of $64 \times 64$.  In the training process, a $64 \times 64$  crop is randomly sampled from the original image  or its  horizontal flip with a padding size of $8$. The initial learning rate is $0.1$ for ResNet18 and  $0.05$  for MobileNetV2 and ShuffleNetV2.

\begin{table}[!h]
\small
  \centering
  \begin{tabular}{cccc}
         \toprule[2pt]
    Model & ResNet18 &  MobileNetV2 & ShuffleNetV2  \\
        \toprule[1pt]

     Baseline  & 53.18 & 49.94 & 46.52\\
     SWA & 54.70 & 48.92 & 47.20 \\
     {SAM} & 55.24 & 51.82 & 47.62 \\
           HWA & \textbf{57.10} & \textbf{53.16} & \textbf{49.34}\\
       \toprule[2pt]
  \end{tabular}
  \caption{  Top-1 Classification  Accuracy   on Tiny-ImageNet.
  }
  \label{tab:tinyimagenet}
\end{table}

\noindent\textbf{CUB200-2011:}  Table~\ref{tab:CUB200} shows the top-1 test  accuracy of different models  on  CUB200-2011, which contains $11788$ images of $200$ categories belonging to birds.  During training, a $224 \times 224$  crop is randomly sampled from the original image or its  horizontal flip and no other data augmentation is applied. The initial learning rate is $0.1$ for all models.

\begin{table}[!h]
\small

  \centering
  \begin{tabular}{cccc}
         \toprule[2pt]
   Model & ResNet18 &  MobileNetV2& ShuffleNetV2  \\
            \toprule[1pt]

     Baseline &58.76&62.43&58.92\\
      SWA&63.41&64.63&60.96\\
      {SAM} &60.31  &63.08 & 59.90 \\
      HWA&\textbf{65.12}&\textbf{69.44}&\textbf{65.44}\\
       \toprule[2pt]
  \end{tabular}
  \caption{  Top-1 Classification Accuracy  on  CUB200-2011.
  }
  \label{tab:CUB200}
\end{table}

\noindent\textbf{CalTech256:} In Table~\ref{tab:caltech256} we report the top-1 test accuracy  on  CalTech256, an object recognition dataset comprising  $30607$ real-world images with different sizes over $256$ object classes and an additional clutter class. The experiment setting for CalTech256 is the same as CUB200-2011 except that the initial learning rate is $0.05$ for all architectures. 
\begin{table}[!h]
\small

  \centering
  \begin{tabular}{cccc}
         \toprule[2pt]
   Model & ResNet18 &  MobileNetV2& ShuffleNetV2  \\
            \toprule[1pt]
        Baseline &64.59&66.30&62.50\\
      SWA &68.30&68.39&64.37\\
       {SAM} & 66.43 & 67.28 & 63.73 \\
            HWA&\textbf{69.33}&\textbf{69.55}&\textbf{67.20}\\

       \toprule[2pt]
  \end{tabular}
  \caption{
  Top-1 Classification Accuracy   on CalTech256.
  }
  \label{tab:caltech256}
\end{table}

Compared to CIFAR, these datasets are more difficult as they have fewer training samples and more classes. As we can observe from Table~\ref{tab:tinyimagenet}, \ref{tab:CUB200} and \ref{tab:caltech256}, our HWA outperforms the baseline and SWA consistently across these datasets and  architectures, which demonstrates the effectiveness of HWA  in these more difficult classification tasks.

\noindent\textbf{ImageNet}  We also evaluate HWA on ImageNet~\cite{krizhevsky2012imagenet}, a large-scale image  dataset with over $1.3$ million images of $1000$ classes. In the experiment, each model is trained with  a batch size of $256$  for $90$ epochs  by  SGD with  a momentum of $0.9$ and a weight decay of $ 1e-4$.   During training, a random resize and crop of $224 \times 224$ and random horizontal flip are performed as data augmentation. For ResNet50, we also apply the cutout regularization~\cite{devries2017improved}  in the training process.

In Table~\ref{tab:imagenet}, we report the top-1 and top-5 test accuracy of each method on ImageNet. As we can observe, HWA outperforms the baseline and SWA by $1.05\%$ and $0.75\%$ for ResNet50, respectively. Therefore, the experimental results  on ImageNet verify that  HWA also works well in the large-scale image classification task.

\begin{table}[h!]
\small
  \centering
  \begin{tabular}{l|cccc}
  \toprule[2pt]
~ &  Baseline & SWA& {SAM}& HWA \\ 
  \toprule[1pt]
  Top-1&76.67&76.97&77.11&77.72\\
  Top-5&93.21&93.41&93.54&93.78\\

  \bottomrule[2pt]
  \end{tabular}
  \vspace{3pt}
  \caption{Top-1 and Top-5 Classification Accuracy on ImageNet.}
  \label{tab:imagenet}
\end{table}

\subsection{Faster Convergence}
\label{sec:expd}
In Section~\ref{sec:discussefficiency}, we show that HWA improves the convergence speed  by performing the weight averaging operation in both horizontal and vertical directions hierarchically. For example, as we can observe in Figure~\ref{fig:VS2} and  \ref{fig:mobv2},  HWA   converges much faster than online WA and offline WA alone.  In Figure~\ref{fig:c100converge}, we provide more  similar experimental results for different architectures (\ie, VGG16, MobileNetV2 and ResNet110) on CIFAR100. For HWA,  the synchronization period is  $H=391$ and the slide window   length is $I=20$. As we can observe, HWA converges  much faster than the baseline with the same optimizer and learning rate scheduler.

\section{Conclusion}
\label{sec:conclusion}
In this work,  we propose a novel training paradigm called hierarchical weight averaging (HWA), which incorporates  the online WA and offline WA into a unified training framework. Unlike online or offline WA, which only serves a single purpose, \ie,  improving training
efficiency or generalization ability, the hybrid paradigm of HWA is able to achieve both of them with better performance. By performing the weight averaging operations in both horizontal and vertical directions hierarchically, HWA outperforms  online WA and offline WA alone. We also analyze how HWA improves the training efficiency and generalization ability of DNNs from the perspective of the loss landscape of DNNs. Finally,  extensive experimental results  demonstrate the effectiveness of the proposed HWA.

\bibliographystyle{IEEEtranN}
\bibliography{egbib}

\end{document}